%% file: main.tex
\documentclass{article}
\usepackage{ijcai21}

\usepackage{times}

\usepackage{soul}
\usepackage{url}
\usepackage[hidelinks]{hyperref}
\usepackage[utf8]{inputenc}
\usepackage[small]{caption}
\usepackage{graphicx}
\usepackage{amsmath}
\usepackage{booktabs}
\usepackage{natbib}
\renewcommand{\cite}{\citep}

\usepackage{amsfonts}
\usepackage{todonotes}
\usepackage{multirow}
\usepackage{bbm}
\usepackage{adjustbox}
\usepackage{subcaption}

\usepackage{tikz}
\usepackage[utf8]{inputenc}
\usepackage{pgfplots}
\DeclareUnicodeCharacter{2212}{−}
\usepgfplotslibrary{groupplots,dateplot}
\usetikzlibrary{patterns,shapes.arrows}
\pgfplotsset{compat=newest}

\makeatletter
\newcommand*\titleheader[1]{\gdef\@titleheader{#1}}
\AtBeginDocument{%
\let\st@red@title\@title
\def\@title{%
\bgroup\normalfont\large\centering\@titleheader\par\egroup
\vskip1.5em\st@red@title}
}
\makeatother

\title{Leveraging Uncertainty for Improved Static Malware Detection Under Extreme False Positive Constraints}

\titleheader{\emph{IJCAI-21 1\textsuperscript{st} International Workshop on Adaptive Cyber Defense}}

\author{
Andre T. Nguyen$^{1,2,3}$\and
Edward Raff$^{1,2,3}$\and
Charles Nicholas$^3$\And
James Holt$^1$\\
\affiliations
$^1$Laboratory for Physical Sciences\\
$^2$Booz Allen Hamilton\\
$^3$University of Maryland, Baltimore County\\
\emails
\{andre,edraff\}@lps.umd.edu,
nicholas@umbc.edu,
holt@lps.umd.edu
}

\begin{document}

\maketitle

\begin{abstract}
The detection of malware is a critical task for the protection of computing environments. This task often requires extremely low false positive rates (FPR) of 0.01\% or even lower, for which modern machine learning has no readily available tools. We introduce the first broad investigation of the use of uncertainty for malware detection across multiple datasets, models, and feature types. We show how ensembling and Bayesian treatments of machine learning methods for static malware detection allow for improved identification of model errors, uncovering of new malware families, and predictive performance under extreme false positive constraints. In particular, we improve the true positive rate (TPR) at an actual realized FPR of 1e-5 from an expected 0.69 for previous methods to 0.80 on the best performing model class on the Sophos industry scale dataset. We additionally demonstrate how previous works have used an evaluation protocol that can lead to misleading results.
\end{abstract}

\section{Introduction}

Classifying a new file as benign (safe to run) or malicious (not safe, termed ``malware'') is a current and growing issue. Malware already causes billions in damages \cite{Anderson2019MeasuringCybercrime,Hyman2013Cybercrime:Serious}, and with healthcare systems increasingly targeted has directly led to deaths~\cite{Eddy2021}. For years most anti-virus (AV) vendors have been seeing at least 2 million malicious new files per month \cite{tagkey2014iv}, and benign files on a network tend to outnumber malicious files at a ratio of 80:1~\cite{Li2017}. This creates a common need for malware detection systems to operate with extremely low false positive rates. If false positives are too frequent, then analysts, IT, and support staff have to spend too much work on non-threats while simultaneously interrupting normal workflow. Even with this focus, Computer Incident Response Teams (CIRT) are often dealing with over 50\% false positive rates and cite it as the main issue with current tooling~\cite{NISC2020}. 

The natural goal for AV style deployments of a malware detector is to maximize the true positive rate (TPR) for some maximally acceptable false positive rate (FPR). Production deployments are often concerned with FPRs of 0.1\% at most, and preferably $\leq$0.001\%. The issue of low FPR has been recognized since the very first research on machine learning based malware detection~\cite{Kephart:1995:BID:1625855.1625983}, yet surprisingly little work has been done to study how to maximize TPR@FPR. We present the first work addressing this gap by applying ideas from ensembling and Bayesian uncertainty estimation to a variety of common malware detection methods in use today on the two largest public corpora.
We develop a number of contributions and previously unrealized insights:

    1) All prior malware detection work we are aware of have evaluated TPR@FPR incorrectly or or did not specify their approach. The common error is to measure the TPR at the desired FPR on the test set, but this presupposes knowledge of the exact threshold to achieve the desired FPR. By instead estimating the threshold on a validation set, we show prior results have often misidentified their true TPR rates. 
    
    2) While the benefits of ensembling have long been known, it is often presumed that significant model diversity is required to obtain meaningful benefit. We show even moderately diverse or Bayesian approaches can significantly improve the TPR, especially for the low-FPR regimes needed for malware detection. 
    
    3) By using a Bayesian approach to estimate the epistemic and aleatoric uncertainty of a model on a given sample, we develop a new approach to thresholding a model's decision that can improve TPR and better approximate the desired FPR on unseen data. 
    
    4) Malware detection deployment requires detecting novel malware families, an intrinsically out-of-distribution task. We show how epistemic and aleatoric uncertainty relates to errors and novel malware families, allowing for faster detection of new threats. 

The rest of this work is organized as follows. First we will review the related research to our work in  \autoref{sec:related_work}. Next we will detail the data, algorithms, and metrics used in all of our experiments in \autoref{sec:methods}. We present extensive experimental results organized by our major contributions in \autoref{sec:results}, which show that prior TPR estimates could be off by $\geq 35\%$ relative error, that ensembles of limited diversity can raise TPR rates by $\geq 11\%$. Then we leverage uncertainty estimates to show a statistically significant improvement over the naive approach of thresholding models for TPR/FPR trade-offs, and that our uncertainty estimates are useful to malware analysts in identifying mispredicted files. Our conclusions are presented in \autoref{sec:conclusion}.

\section{Related Work} \label{sec:related_work}

The need for low FP rates has been paramount since the inception of machine learning malware detection research by \citet{Kephart:1995:BID:1625855.1625983}. Much of the history in this domain is focused on ``signature'' like tasks, where the goal was to recognize the set of already known malware, smaller than the total population of malware. This led to works that used \textit{the training data as part of the final evaluation data}~\cite{Gavrilut2012,Fukushima2010}. This approach is not meaningful for determining TPR at any FPR due to over-fitting, and is not tenable due to the now large and growing population of malware with more sophisticated obfuscation techniques. There is no agreed-upon threshold for exactly how low FPs should be, with most published work ranging from 0.1\% down to $\leq$0.00002\% \cite{Rafique:2013:FMC:2941590.2941601,smutz2012malicious,Pitsillidis2010,Anderson2016,Perdisci2008,182795,Saxe2015,Kilograms_2019}. Beyond some works using training data at test time, all of these works \textit{evaluate their false positive rates on the test set, selecting the threshold from the test set that gives them the desired FPR}, and then report the associated TPR. This is an understandable but incorrect approach, because the threshold is selected explicitly from the test set, when our goal is to test the ability of the model to achieve an FPR on unseen data. As far as we are aware, our work is the first in the malware detection space to identify this and propose selecting the threshold from a validation set, and then evaluate the precision of the FPR estimate in conjunction with the obtained TPR.

It is also worth noting that these cited prior works attempt to minimize FPR primarily by feature selection, engineering, or ML process pipeline choices that they believe will result in a more accurate model or be biased towards low FPR. Our approach is instead model agnostic, and seeks to better understand the nature of selecting thresholds to achieve TPR@FPR targets and improving it with uncertainty estimates. The only other work we are aware of that has this goal is from the related domain of spam detection by \citet{learning-at-low-false-positive-rates} who propose two dataset re-weighting approaches, but also determine their success using the test set. Because they focus on data re-weighting their approach is orthogonal to our own, and both could be used simultaneously. The closest other work we are aware of is \cite{pmlr-v54-eban17a} that develop differentiable approximations to AUC and Precision at a minimum Recall, but their approach does not apply to our situation because we instead need a maximum FP rate and specific points on the AUC curve. Further,  our need for very low FPR is problematic for their setting as a mini-batch of data will be insufficient for estimating low FPR. 

A number of prior works have investigated diverse ensembles of different kinds of algorithms to improve malware detection accuracy \cite{Ye:2010:AMC:1835804.1835820,Liu2016a,Kang:2016:EMD:2835776.2835834,Menahem:2009:IMD:1497631.1497723,Khasawneh2015}, following the common wisdom that averaging over diverse and uncorrelated predictors improves accuracy~\cite{Wolpert1992,Breiman1996,Jacobs1991}. As far as we are aware, we are the first to study the performance of small ensembles of low-diversity (i.e, different runs of the same algorithm) and identify their especially large impact on TPR when needing extremely low FPR. This is important for malware detection as a diverse ensemble often necessitates algorithms that are too slow for deployment, and high compute throughput is critical to practical utility in this domain.  

Building upon the use of ensembles, the modeling of uncertainty for decision making is notably missing from the current machine learning for malware literature. An exception is the approach of \citet{Backes2017LUNA:Learning} which introduces new classes for uncertain decisions in the context of a simple Bayesian logistic regression model applied to the Drebin Android malware dataset \cite{Arp2013TechnicalPocket,Spreitzenbarth2013Mobile-sandbox:Applications}. Understanding when a machine learning model is uncertain about its prediction is critical in high risk applications such as malware detection. When an automated malware detection algorithm is uncertain about a sample, the uncertainty estimate could be used to flag the sample for analysis by a more computationally expensive algorithm or for review by a human. Our work is the first we are aware to study how modeling uncertainty can be used to improve TPR@FPR scores, and to aid analysts by showing new samples with high uncertainty are more likely to be novel malware families.

\section{Methods} \label{sec:methods}

We provide details about the data and machine learning models used in our experiments. The majority of the existing research in machine learning applied to malware detection has focused on the automation of static malware analysis, where a file is analyzed without being run. We will also focus on the static analysis problem.

\subsection{Data}

Due to the need to estimate low FPR rates, we use the two largest available corpora for malware detection. These are the EMBER2018 and Sophos AI SOREL-20M datasets. We note that both of these datasets focus on low FPR evaluation, but make the same error in evaluation. Our first results in \autoref{sec:methods} will show the relative magnitudes of the errors. 

We use the EMBER2018 dataset which consists of portable executable files (PE files) scanned by VirusTotal in or before 2018 \cite{Anderson2018EMBER:Models}. The dataset contains 600,000 labeled training samples and 200,000 labeled testing samples, with an equal number of malicious and benign samples in both sets. The malicious samples are also labeled by malware family using AVClass \cite{SebastianAVCLASS:Labeling}. All of the testing samples were first observed after all of the training samples. EMBER2018 includes vectorized features for each sample encoding general file information, header information, imported functions, exported functions, section information, byte histograms, byte-entropy histograms, and string information \cite{Anderson2018EMBER:Models}. While the 1.1TB of raw PE files are not available as part of EMBER2018, they can be downloaded via VirusTotal. We note that the EMBER2018 dataset was designed to be more challenging for machine learning algorithms to classify than the original EMBER2017 dataset.

We also use the recent Sophos AI SOREL-20M dataset, consisting of 20 million files \cite{Harang2020SOREL-20M:Detection}. The much larger number of data points in the Sophos dataset is advertised as ``industry scale'' and allows for the exploration of FPR constraints much smaller than allowed by EMBER2018. In particular, the test set size for Sophos consists of 1,360,622 malicious samples and 2,834,441 benign samples. As part of the Sophos dataset release, two baseline models are provided, including a feed-forward neural network (FFNN) and a LightGBM (LGBM) gradient-boosed decision tree model. Five versions of each of the models are pre-trained using different random seeds on the Sophos data using the same featurization as EMBER2018. 

\subsection{Models}

\textbf{EMBER2018:} We apply three models to the EMBER2018 dataset that each rely on different types of features. 
The first model we apply is a Bayesian deep learning model based on the MalConv model of \citet{Raff2017MalwareEXE}, a convolutional neural network for malware detection that operates on the raw byte sequences of files. We will refer to this model as Bayesian MalConv (BMC). As exact Bayesian inference cannot be done for deep neural networks such as MalConv, approximate inference methods need to be used. 

\citet{Gal2016DropoutLearning} introduced an easy to implement approach to variational inference in Bayesian neural networks. In particular, they showed that a neural network with dropout, a technique commonly used to reduce overfitting in neural networks by randomly dropping units during training \cite{Hinton2012ImprovingDetectors,Srivastava2014Dropout:Overfitting}, applied before every weight layer is equivalent to an approximation of a deep Gaussian process \cite{Damianou2013DeepProcesses}, and that training with dropout effectively performs variational inference for the deep Gaussian process model. The posterior distribution can be sampled from by leaving dropout on at test time. 
For Bayesian MalConv, we follow this approach and apply dropout before each fully connected layer of the MalConv model, with a dropout probability of $p=0.1$. We use the Adam optimizer \cite{Kingma2014Adam:Optimization} to train the model, and we produce 16 samples at evaluation time using multiple forward passes on the trained model with dropout left on.  

The second model we apply is a Bayesian logistic regression (BLR) model which takes as input the binary presence of 94,225 byte 8-grams \cite{Kilograms_2019,Raff2018AnClassification} selected using LASSO from the one million most common byte 8-grams. Dropout is used in a similar manner to Bayesian MalConv, with a dropout probability of $p=0.1$ and 16 samples at evaluation time.

The third model we apply is an ensemble of 16 Light Gradient Boosting Machine (LGBM) models \cite{KeLightGBM:Tree} trained with different random seeds on the EMBER features as described in \citet{Anderson2018EMBER:Models}.


\textbf{Sophos:} We apply two models to the Sophos dataset that both rely on the EMBER2018 featurization. The first is an ensemble of 5 feed-forward neural network (FFNN) models as described in \citet{Harang2020SOREL-20M:Detection}, a simplified version of the model from \citet{RuddALOHA:Augmentation}, trained using different random seeds. The second is an ensemble of 5 LGBM models trained using different random seeds. We use the publicly available pre-trained models provided with the Sophos dataset as our ensemble members for both models. 
While 5 ensemble members may seem small, \citet{Ovadia2019CanShift} found that increasing ensemble sizes beyond 5 has diminishing returns with respect to the quality of the uncertainty estimates, so an ensemble size of 5 may be sufficient. Our results will show that not only are they sufficient, but for our goal of low FPR they can be significantly more effective than has been previously reported. 

\subsection{Uncertainty Estimation}

The Bayesian framework allows for the principled modeling of uncertainty in machine learning and decision making. Within this framework, probabilities represent degrees of belief as opposed to the frequentist interpretation of probabilities as long run frequencies \cite{Liu2014BayesianInference}. Bayesian inference uses Bayes' Theorem to update beliefs (that are represented in the form of probability distributions) when new data is observed.

In the context of machine learning, a Bayesian update takes the following form where $\theta$ represents model parameters, $D$ represents the data, and $M$ represents the model class: 
$\mathbb{P}(\theta|D,M) = \frac{\mathbb{P}(D|\theta,M)\mathbb{P}(\theta|M)}{\mathbb{P}(D|M)}$, where 
$\mathbb{P}(\theta|D,M)$ is the posterior belief about the model parameters given the data, $\mathbb{P}(D|\theta,M)$ is the likelihood of the data given the model parameters, $\mathbb{P}(\theta|M)$ is the prior belief about model parameters, and $\mathbb{P}(D|M)$ is the marginal likelihood or evidence. Bayesian inference is usually intractable due to the integrals involved, unless the prior distribution is conjugate to the likelihood distribution. Unfortunately, conjugate priors exist for only exponential family distributions \cite{Murphy2012MachineSeries} and so can't be directly applied to complex models like Bayesian deep neural networks. 

As exact Bayesian inference cannot be done for Bayesian deep learning models, approximate inference methods need to be used. Given sufficient compute time, Markov Chain Monte Carlo (MCMC) methods can be used to sample from the posterior \cite{Neal1995BayesianNetworks}. Unfortunately, common sampling based approaches are difficult to scale to problems in the malware space where practical dataset sizes are measured in terabytes because they require gradient computations over the entire dataset. As MCMC is hard to scale in practice, variational inference is often used instead which converts the integration problem into an optimization problem where the posterior is approximated using a simpler variational distribution \cite{Blei2017VariationalStatisticians}. Variational inference for neural networks was first introduced in the early nineteen nineties \cite{Hinton1993KeepingWeights}, and \citet{Graves2011PracticalNetworks} revived interest in variational inference for neural networks by introducing a stochastic variational method for inference in neural networks. 

We note that complicated Bayesian inference is not necessarily needed to provide useful uncertainty estimates. \citet{Lakshminarayanan2017SimpleEnsembles} introduce an alternative that trains an ensemble of randomly initialized models. These deep ensembles have been shown to produce competitive uncertainty estimates \cite{Ovadia2019CanShift,Ashukha2020PitfallsLearning} because they are able to explore different modes in function space \cite{Fort2019DeepPerspective}. \citet{Wilson2020BayesianGeneralization} argue that deep ensembles are not a competing approach to Bayesian deep learning but rather are an effective approach for Bayesian model averaging. 

Two kinds of uncertainty can be distinguished \cite{Gal2016UncertaintyLearning}. Aleatoric uncertainty is caused by inherent noise and stochasticity in the data. More training data will not help to reduce this kind of uncertainty. Epistemic uncertainty on the other hand is caused by a lack of similar training data. In regions lacking training data, different model parameter settings that produce diverse or potentially conflicting predictions can be comparably likely under the posterior.  

For classification tasks where epistemic and aleatoric uncertainty don't need to be differentiated, uncertainty can be measured using the predictive distribution entropy:
$$H[\mathbb{P}(y|x,D)] = - \sum_{y \in C}{\mathbb{P}(y|x,D) \log \mathbb{P}(y|x,D)}$$

\noindent Aleatoric uncertainty 
is
measured using expected entropy: $$u_\mathit{alea} = \mathbb{E}_{\mathbb{P}(\theta|D)}H[\mathbb{P}(y|x,\theta)]$$

\noindent Mutual information
is
used to measure epistemic uncertainty:
$$u_\mathit{epis} = I(\theta,y|D,x) = H[\mathbb{P}(y|x,D)] - \mathbb{E}_{\mathbb{P}(\theta|D)}H[\mathbb{P}(y|x,\theta)]$$

Monte Carlo estimates obtained by sampling from the posterior can be used to approximate the terms of these equations for our Bayesian models \cite{Smith2018UnderstandingDetection}. In particular, $\mathbb{P}(y|x,D) \approx \frac{1}{T} \sum_{i=1}^T \mathbb{P}(y|x,\theta_i)$ and $\mathbb{E}_{\mathbb{P}(\theta|D)}H[\mathbb{P}(y|x,\theta)] \approx \frac{1}{T} \sum_{i=1}^T H[\mathbb{P}(y|x,\theta_i)]$ where the $\theta_i$ are samples from the posterior over models and $T$ is the number of samples. 

For our ensemble based models which are not explicitly Bayesian (because each ensemble member receives the same weight) but Bayesian inspired, uncertainties can be computed in a similar way where the $\theta_i$ are no longer samples from a posterior, but instead multiple independent trainings of a model with $T$ different random seeds.

\subsection{Classification Metrics}

We use multiple metrics to evaluate and compare approaches.

Accuracy is defined as the percent of correct predictions made. Area under the receiver operating characteristic curve (AUC) is the probability that the classifier will rank a randomly selected malicious file higher in probability to be malicious than a randomly selected benign file. The true positive rate (TPR) is defined as the number of true positives over the sum of true positives and false negatives. The false positive rate (FPR) is defined as the number of false positives over the sum of false positives and true negatives. 

An important contribution of our work is to recognize that the TPR obtained at any given FPR on the test set is not the actual measure of interest in malware detection, but an over-fit measure due to the implicit assumption that the correct decision threshold is known at test time. The threshold must be estimated during training or validation, and then applied to the test set. This means we have a target maximum FPR $T_\mathit{FPR}$ that we wish to obtain, and a separate \textit{actualized FPR} that is obtained on the test set.  
In order to capture the trade-off between TPR and actualized FPR constraint satisfaction, we define the following combined metric \autoref{eq:tpr_fpr_trade}  where $T_\mathit{FPR}$ is the desired maximum FPR. 
\begin{equation} \label{eq:tpr_fpr_trade}
    C = TPR-\frac{\max(\text{actualized }FPR-T_\mathit{FPR},0)}{T_\mathit{FPR}}
\end{equation}

This metric captures that we have a desired TPR, but penalizes the score based on the degree of violation of the FPR. This is done by a division so that the magnitude of the violation's impact grows in proportion to the target FPR shrinking. This matches the nature of desiring low FPR itself. For example, 90\% TPR at a target FPR of 0.1\% is still quite good if the actualized FPR is 0.11\% ($C=0.8$), but is unacceptably bad if the target FPR was 0.01\% ($C=-9.1$). 

\section{Experiments and Discussion} \label{sec:results}

Now that we have discussed the methods of our work and the metrics by which they will be examined, we will show empirical results demonstrating our primary contributions: 1) Evaluating test-set performance thresholds from the test set leads to misleading results at lower FPR, 2) Simple non-diverse ensembles can dramatically improve TPR at any given FPR rate, 3) we can further improve TPR@FPR by explicitly modeling Bayesian uncertainty estimates into our decision process, and 4) these uncertainty estimates have practical benefits to application by showing that errors and previously unseen malware families have uncertainty distributions that place more weight on higher uncertainties. For each of these we will include the empirical results on the EMBER2018 and the Sophos 2020 corpora, and include additional discussion and nuance to how these relate to practical deployment. 

\subsection{Misleading Evaluation}

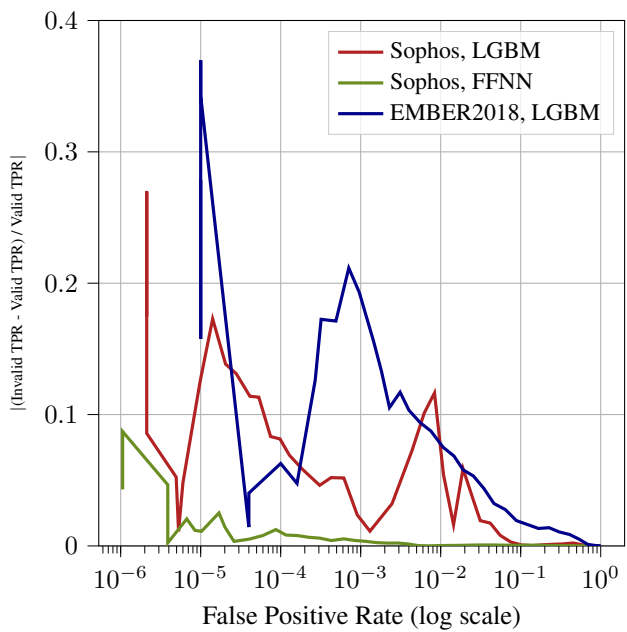
\begin{figure}[!tb]
\centering
\vspace{0ex}
\input{figs/in_valid.tex}
\vspace{-1ex}
\caption{Absolute relative error in TPR when using the invalid evaluation protocol, for three different model and dataset combinations. A valid evaluation protocol will use a validation set ROC curve to select a threshold given a desired FPR. The evaluation protocol that is currently the norm in the malware detection literature is invalid because it uses the test set ROC curve, which is never available in practice, to select a threshold. The use of the invalid evaluation protocol can lead to over a 30 percent relative error in TPR.}
\label{fig:figs/in_valid.pdf}
\end{figure}

A currently accepted practice for evaluating malware detection models under FPR constraints is to report the test set ROC curve. Once the test set ROC curve is produced, the desired FPR rates from the curve are selected to show their associated TPR. This is misleading as in practice the test set is not available when choosing the decision threshold, causing this evaluation procedure to be invalid. 
Instead, we must recognize that there are a priori \textit{target FPRs} that are the FP rates that we desire from the model, and the \textit{actualized FPRs} which are what is obtained on the test (read, ``production'') data. Selecting the threshold from the test set hides that the target and actualized FPRs are different, especially for low FPRs that require large amounts of data to estimate. 
The valid approach to this scenario when evaluating a classifier at different FPRs is to select the thresholds using a validation set. Once the thresholds are selected that obtain the target FPRs, they can be applied to the test set to obtain the actualized FPRs and their associated TPRs. We show the impact this has on the entire TPR/FPR curve in \autoref{fig:figs/in_valid.pdf} which shows the absolute relative error in TPR for a given actualized FPR. Depending on the model and dataset, the resulting TPR for any actualized FPR can change by over 30\%, and the relative error generally increases as the FPR decreases. This is expected because low FPRs naturally require more data to estimate: if you want an FPR of 1:1,000 and you want 1,000 FPRs to estimate the threshold from you would expect to need $1,000^2 = $ 1 million examples.

\begin{figure}[!tb]
\centering
\vspace{0ex}
\input{figs/subsample.tex}
\vspace{-1ex}
\caption{Absolute relative error in TPR when using the invalid evaluation protocol at various levels of subsampling of the validation set.}
\label{fig:figs/subsample.pdf}
\end{figure}
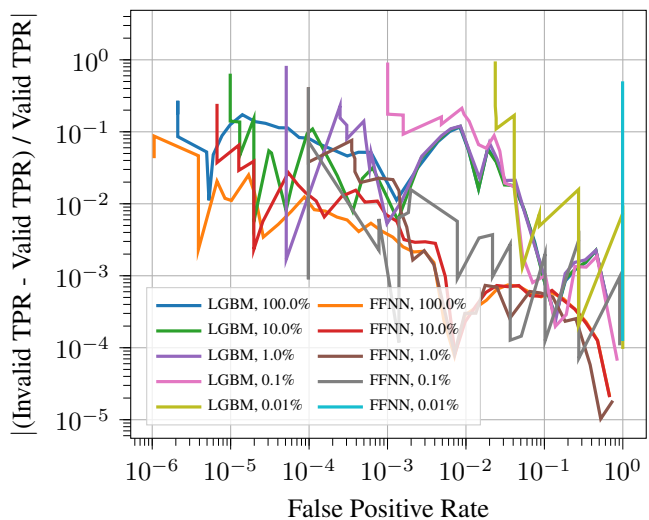

We note that the Sophos FFNN model seems to be particularly robust with the lowest error in \autoref{fig:figs/in_valid.pdf}. This is in part a testament to the FFNN approach, but more broadly a function on the magnitude of the Sophos dataset. With 2.5 million samples in the validation set and 4.2 million in the test set, the corpus is large enough to mitigate the impact of some inappropriate practices. To demonstrate the impact the validation set can have, we show the same results in \autoref{fig:figs/subsample.pdf} when only the validation set used to select the threshold is reduced by various orders of magnitude. 

One can clearly see that as the validation set size decreases, the ability to estimate the FPR decreases. This causes more errors and a ``shortening'' of the curves as it becomes impossible to estimate lower desired FPR rates. This last point is important as some prior works have reported FPRs lower than what their dataset could accurately estimate. If the test set size times the desired FPR is less than 100 samples, it is unlikely the TPR@FPR reported will be an accurate estimate (e.g., as done in \cite{Anderson2016}). 

We note that this distinction between invalid vs. valid approaches is \textit{not} a critique on the evaluation of entire ROC curves. The fundamental distinction is whether we care about the entire ROC curve, or only specific points from the ROC curve. If you care about the entire ROC curve, evaluating the ROC on the test set is valid and appropriate. But because malware detection is concerned with particular points from the ROC curve, it becomes necessary to evaluate if the approach can hit its desired location on the curve (i.e., a specific FPR in production). There are also valid scenarios to consider just the ROC curve as a whole for malware analysis and its associated AUC, as it represents a metric of ability to rank that is applicable to other scenarios within malware detection and analysis. Our critique is for just those concerned with AV-like deployments that aim for low FPRs specifically.

\subsection{Ensembles}

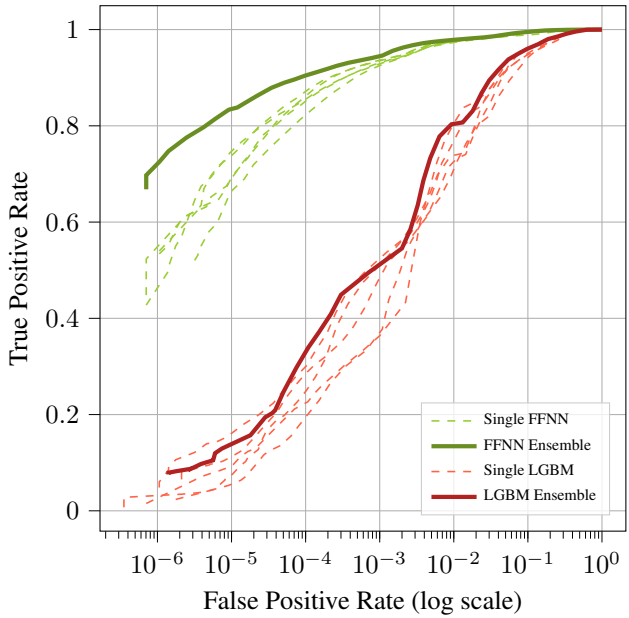
\begin{figure}[!tb]
\centering
\vspace{0ex}
\input{figs/ensemble_sophos.tex}
\vspace{-1ex}
\caption{Sophos ensembles that take the average of predictions from randomly seeded models can lead to significant TPR gains under extreme FPR constraints, compared to individual models.}
\label{fig:figs/ensemble_sophos.pdf}
\end{figure}

We have now shown that the correct approach to developing a ROC curve when one wishes to evaluate specific points on the curve is to select the threshold from a validation set rather than the test set. We will apply this to the results of this section to show that creating an ensemble of randomly seeded models can improve the obtained TPR at almost any actualized FPR, especially under extreme FPR constraints. 
 \autoref{fig:figs/ensemble_sophos.pdf} shows the ROC curves for individual models as well as for the ensemble consisting of those individual models. The Sophos trained FFNN ensemble notably performs significantly better than any individual member of the ensemble, with the gap in performance widening as FPR becomes smaller. 
 
The fact that these are all the same type of model, but with different random seeds at initialization, is an important and not previously recognized phenomena. Classical wisdom is that ensembles should maximize diversity to reduce correlation of predictions, and thus maximize accuracy. But in our scenario malware detection models are designed to be lightweight in model size, low latency, and high throughput, so that the AV system does not interrupt the users of a computing system. A classically diverse model with different types of algorithms or features, as done in prior work in this space, ends up including approaches that are many orders of magnitude larger and slower than the lighter weight approaches we study in this work. Because we can use multiple versions of the same type of model with the same features, we can maintain the high throughput, low latency \& size requirements while obtaining these large gains in TPR. 

\autoref{tab:accs} compares the accuracy, AUC, and AUC $@ \leq$ 0.1\% FPR achieved by ensembles to the average of those achieved by individual ensemble members. In all cases, the ensemble has better performance than the expected performance of individual ensemble members, even though they are using an ensemble of low diversity. Of particular importance is the performance of each model in the low FPR domain has a greater relative improvement (median 11\% improvement) than one may have anticipated looking at more standard metrics like Accuracy and AUC (median improvements of 0.6\% and 0.4\% respectively). The only exception to this is the Logistic Regression approaches which have difficulty operating at the extremely low FPR ranges, which we will see repeated. 

\begin{table}[!tb]
\caption{Accuracy and AUC for each model and dataset combination. Ensembles are compared to the expected performance of their components. Best results are shown in \textbf{bold}.}
\adjustbox{max width=\columnwidth}{
\begin{tabular}{lllll}
\toprule
\multicolumn{1}{c}{Dataset} & \multicolumn{1}{c}{Model} & \multicolumn{1}{c}{Accuracy} & \multicolumn{1}{c}{AUC} & \multicolumn{1}{c}{$\text{AUC}_{\leq 0.1\% FPR}$} \\ \midrule
\multirow{6}{*}{EMBER}      & Bayesian MalConv          & \textbf{91.64}               & \textbf{97.47}          & \textbf{0.04079}                           \\
                            & MalConv                   & 90.88                        & 97.05                   & 0.03288                                    \\
                            & Bayesian Log. Reg.        & \textbf{94.72}               & \textbf{98.15}          & \textbf{0.0}                               \\
                            & Log. Reg.                 & 94.15                        & 97.32                   & \textbf{0.0}                               \\
                            & LightGBM Ensemble         & \textbf{93.98}               & \textbf{98.62}          & \textbf{0.06054}                           \\
                            & LightGBM                  & 93.88                        & 98.55                   & 0.05433                                    \\ \midrule
\multirow{4}{*}{Sophos}     & FFNN Ensemble             & \textbf{98.81}               & \textbf{99.83}          & \textbf{0.09274}                           \\
                            & FFNN                      & 98.56                        & 99.75                   & 0.08990                                    \\
                            & LightGBM Ensemble         & \textbf{86.10}               & \textbf{98.41}          & \textbf{0.04459}                           \\
                            & LightGBM                  & 85.47                        & 98.05                   & 0.03637                                    \\ \bottomrule
\end{tabular}
}
\label{tab:accs}
\end{table}

\subsection{Uncertainty Based Threshold Adjustments}

While ensembling predictions by taking an average leads to improved results, there is more information within ensembles that can be leveraged to further ameliorate performance under extreme FPR constraints. In particular, estimates can be computed for epistemic ($u_\mathit{epis}$) and aleatoric ($u_\mathit{alea}$) uncertainty. We introduce a simple threshold adjustment approach that leverages data point specific uncertainty to locally adjust the threshold based on the uncertainty. We explore three uncertainty aware local threshold adjustments:

\begin{align}
  \hat{y}_\mathit{adj} &= \hat{y} + \alpha_1 \cdot u_\mathit{epis} + \alpha_2 \cdot u_\mathit{alea} \label{eq:lv1} \\
  \hat{y}_\mathit{adj} &= \hat{y} + \alpha_1 \cdot \exp (\alpha_3 \cdot u_\mathit{epis}) + \alpha_2 \cdot \exp (\alpha_4 \cdot u_\mathit{alea}) \label{eq:lv2} \\
\begin{split}
    \hat{y}_\mathit{adj} &= \hat{y} + \mathbbm{1}[\hat{y}>\alpha_0] (\alpha_1 \cdot u_\mathit{epis} + \alpha_2 \cdot u_\mathit{alea}) +\label{eq:lv3} \\
    &\quad \mathbbm{1}[\hat{y} \leq \alpha_0] (\alpha_3 \cdot u_\mathit{epis} + \alpha_4 \cdot u_\mathit{alea})
   \end{split}
\end{align}

\noindent where $\mathbbm{1}[\cdot]$ is the indicator function, $\hat{y}$ is the original ensemble prediction for a data point,  $u_\mathit{epis}$ is the epistemic uncertainty (mutual information) for a data point's prediction, $u_\mathit{alea}$ is the aleatoric uncertainty (expected entropy) for a data point's prediction, the $\alpha_i$ are learned scaling factors, and $\hat{y}_\mathit{adj}$ is the uncertainty adjusted prediction for a data point. The scaling factors are learned by iteratively optimizing each $\alpha_i$ to maximize TPR given a desired FPR, where after each scaling factor adjustment a new global adjustment is computed. 

Because TPR@FPR is not a differentiable metric, we use a gradient free approach to altering the weights $\boldsymbol{\alpha}$. In particular, we use a coordinate descent style approach where we take turns optimizing the individual $\alpha_i$ values while holding all others fixed, repeating the process until convergence. This is feasible thanks to the convex behavior that occurs with respect to the TPR scores. If $\alpha_i$ is set too large, then the associated feature (e.g, $u_\mathit{epis}$ or $u_\mathit{alea}$) becomes the only effective factor by overshadowing all other components, but is not sufficient on its own to make meaningful predictions, resulting in a low TPR when selecting the target FPR. If the weight is too small then the associated feature has no impact, and the result is unchanged. This creates the two ``low points,'' and a weight that results in a higher score (hopefully) exists between the two extrema. This can then be selected with high precision by using a golden search by treating the extrema as brackets on a solution. We use Brent's method to solve this because it allows faster searches by approximating the problem with a parabola when possible, and switching to golden search in the worst case, allowing it to solve the optimization quickly.\footnote{In our experience each call to Brent's method takes less than 100 optimization steps.}

This gives us an approach to directly optimize our target metric even though it is non-differentiable, and to do so with high precision in just a few minutes of computation. All optimization occurs on the validation set. While the Sophos dataset comes with a validation set, EMBER2018 does not, so we create a validation set for EMBER2018 using five fold cross-validation.  
The $\alpha_i$ are optimized in an alternating manner for \autoref{eq:lv1} and in a randomized sequential manner for \autoref{eq:lv2} and \autoref{eq:lv3}. Brent's method is used as the optimizer \cite{Brent1972AlgorithmsDerivatives}, with a bracketing interval of $[-100,100]$ for \autoref{eq:lv1}, of $[-10,10]$ for \autoref{eq:lv2}, and of $[0,1]$ for \autoref{eq:lv3} with $[-0.1,0.1]$ for $\alpha_0$. A ROC curve is then computed using the locally adjusted $\hat{y}_\mathit{adj}$ to obtain the final global threshold. For all methods, when fitting the global and local threshold adjustments using the validation set, a target FPR of $0.9$ times the actual desired FPR is used in order to conservatively satisfy the constraint given differences in the test set data.

We briefly note that we had also tried optimizing the uncertainty scaling factors and thresholds jointly using a CMA-ES gradient-free optimization approach \cite{HansenCompletelyStrategies}. Unfortunately, the obtained solutions were not precise enough given the extremely small FPR constraints, leading us to use the iterative optimization of each variable using Brent's method.

The results comparing our new uncertainty augmented local adjustments \autoref{eq:lv1} (g+l) and \autoref{eq:lv2} (g+lv2) and \autoref{eq:lv3} (g+lv3) against the naive approach (g) are provided in \autoref{tab:sophos-val} for the Sophos dataset. Additional results for EMBER2018 are provided in Appendix \autoref{tab:ember-val}. Bolded are the methods that performed best by the combined score \autoref{eq:tpr_fpr_trade} which penalizes going over the target FPR. 
We note that across datasets,  models, and target constraints, the inclusion of uncertainty based local adjustments (g+l and g+lv2 and g+lv3) improves over the standard use of only a global threshold adjustment (g). All three approaches are statistically significant in their improvement (Wilcoxon-signed rank test p-values of 0.02, 0.01, and 0.025 respectively).  
In almost all cases if (g+lv2) is the best performer, (g+l) and (g+lv3) are the second and third best. Similarly, when (g+lv2) is not the best, it is usually still better than (g). 

\begin{table}[!tb]
\centering
\caption{Sophos comparison of the standard global adjustment (labeled as (g)) to the uncertainty aware local adjustments from \autoref{eq:lv1} (labeled as (g+l)) and \autoref{eq:lv2} (labeled as (g+lv2)) and \autoref{eq:lv3} (labeled as (g+lv3)). Best combined score (TPR penalized for over-runs on target FPR) shown in \textbf{bold}.}
\label{tab:sophos-val}
\adjustbox{max width=\columnwidth}{%

\begin{tabular}{@{}lrrrrrrrrr@{}}
\toprule
\multicolumn{1}{c}{\multirow{2}{*}{\begin{tabular}[c]{@{}c@{}}Target\\ FPR\end{tabular}}} & \multicolumn{1}{c}{\multirow{2}{*}{\begin{tabular}[c]{@{}c@{}}Test \\ Perf.\end{tabular}}} & \multicolumn{4}{c}{LGBM}                                                          & \multicolumn{4}{c}{FFNN}                                                          \\ \cmidrule(l){3-6}  \cmidrule(l){7-10} 
\multicolumn{1}{c}{}                                                                      & \multicolumn{1}{c}{}                                                                       & \multicolumn{1}{c}{(g)} & \multicolumn{1}{c}{(g+l)} & \multicolumn{1}{c}{(g+lv2)} & \multicolumn{1}{c}{(g+lv3)} & \multicolumn{1}{c}{(g)} & \multicolumn{1}{c}{(g+l)} & \multicolumn{1}{c}{(g+lv2)} & \multicolumn{1}{c}{(g+lv3)}  \\ \midrule
\multirow{3}{*}{1\%}                                                                      & TPR                                                                                        & 8.060E-01               & 8.125E-01                 & \textbf{8.150E-01}          & 8.137E-01 & \textbf{9.779E-01}      & \textbf{9.779E-01}        & \textbf{9.779E-01}    & \textbf{9.779E-01}
      \\
                                                                                          & FPR                                                                                        & 1.175E-02               & 1.123E-02                 & \textbf{1.125E-02}  & 1.129E-02
       & \textbf{8.664E-03}      & \textbf{8.663E-03}        & \textbf{8.666E-03}    & \textbf{8.665E-03}
      \\
                                                                                          & Comb.                                                                                      & 6.315E-01               & 6.899E-01                 & \textbf{6.904E-01}         & 6.845E-01
 & \textbf{9.779E-01}      & \textbf{9.779E-01}        & \textbf{9.779E-01}    & \textbf{9.779E-01}
      \\ \midrule
\multirow{3}{*}{0.1\%}                                                                    & TPR                                                                                        & 5.264E-01               & 5.318E-01                 & \textbf{5.343E-01}      & 5.342E-01
    & 9.440E-01               & \textbf{9.471E-01}        & 9.450E-01 & 9.450E-01
                   \\
                                                                                          & FPR                                                                                        & 1.493E-03               & 1.088E-03                 & \textbf{9.699E-04}    & 9.681E-04
      & 9.695E-04               & \textbf{9.473E-04}        & 1.024E-03           & 1.022E-03
        \\
                                                                                          & Comb.                                                                                      & 3.338E-02               & 4.434E-01                 & \textbf{5.343E-01}      & 5.342E-01
    & 9.440E-01               & \textbf{9.471E-01}        & 9.208E-01 & 9.233E-01
                   \\ \midrule
\multirow{3}{*}{0.01\%}                                                                   & TPR                                                                                        & 2.296E-01               & 2.352E-01                 & \textbf{2.371E-01}       & 2.278E-01
   & 9.017E-01               & 9.037E-01                 & \textbf{9.086E-01}    & \textbf{9.086E-01}
      \\
                                                                                          & FPR                                                                                        & 4.339E-05               & 5.751E-05                 & \textbf{5.786E-05}      & 4.375E-05
    & 8.855E-05               & 9.032E-05                 & \textbf{8.961E-05}   & \textbf{9.102E-05}
       \\
                                                                                          & Comb.                                                                                      & 2.296E-01               & 2.352E-01                 & \textbf{2.371E-01}   & 2.278E-01
       & 9.017E-01               & 9.037E-01                 & \textbf{9.086E-01}    & \textbf{9.086E-01}
      \\ \midrule
\multirow{3}{*}{0.001\%}                                                                  & TPR                                                                                        & 9.940E-02               & 1.007E-01                 & \textbf{1.075E-01}     & 1.007E-01
     & 8.022E-01               & \textbf{8.046E-01}        & 8.043E-01       & 8.041E-01
            \\
                                                                                          & FPR                                                                                        & 3.881E-06               & 4.586E-06                 & \textbf{4.939E-06}  & 4.586E-06
        & 4.234E-06               & \textbf{5.292E-06}        & 5.998E-06           & 4.586E-06
        \\
                                                                                          & Comb.                                                                                      & 9.940E-02               & 1.007E-01                 & \textbf{1.075E-01}     & 1.007E-01
     & 8.022E-01               & \textbf{8.046E-01}        & 8.043E-01          & 8.041E-01
         \\ \bottomrule
\end{tabular}
}
\end{table}

\subsection{Uncertainty on Errors and New AV Classes}

\begin{figure}[!tb]
    \centering
    \begin{subfigure}[t]{\columnwidth}
        \centering
        \includegraphics[width=\columnwidth]{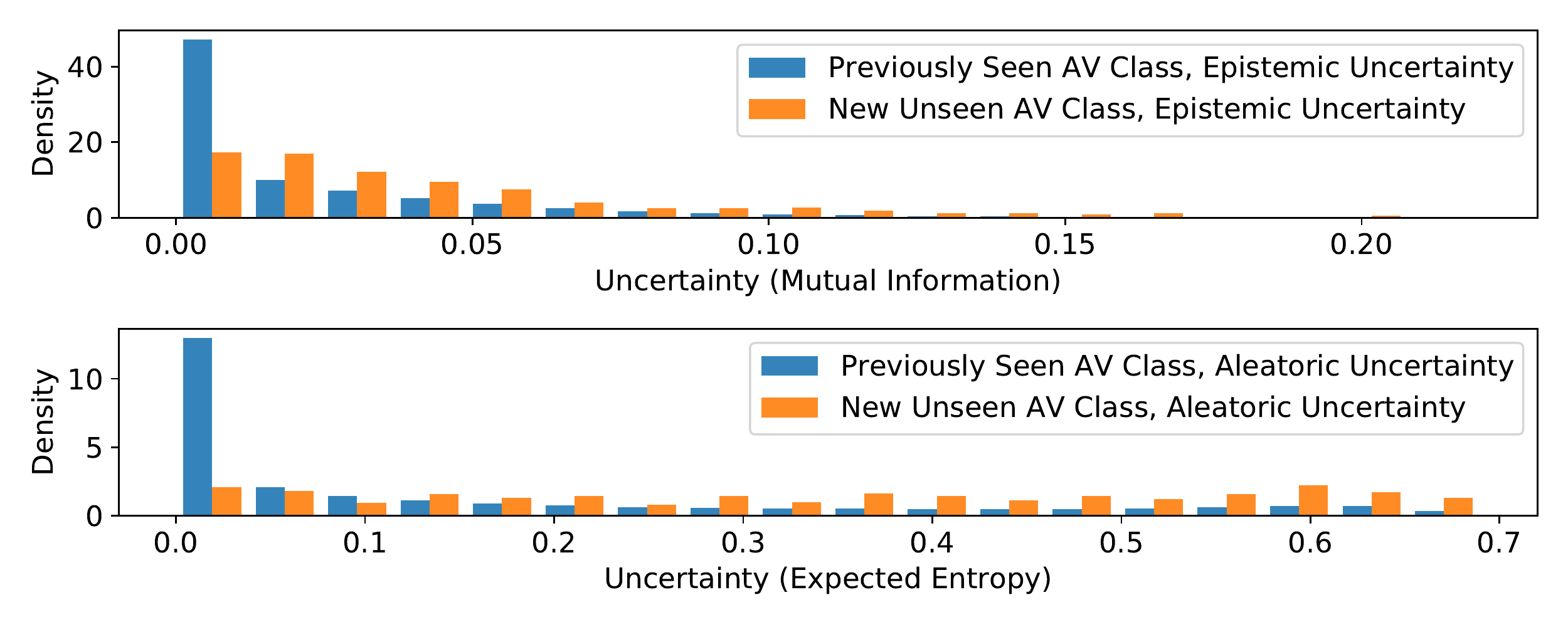}
        \caption{Bayesian MalConv} \label{fig:figs/unseen_ember2018_v1_test_bmc.pdf}
    \end{subfigure}%
    \\
    \begin{subfigure}[t]{1\columnwidth}
        \centering
        \includegraphics[width=\columnwidth]{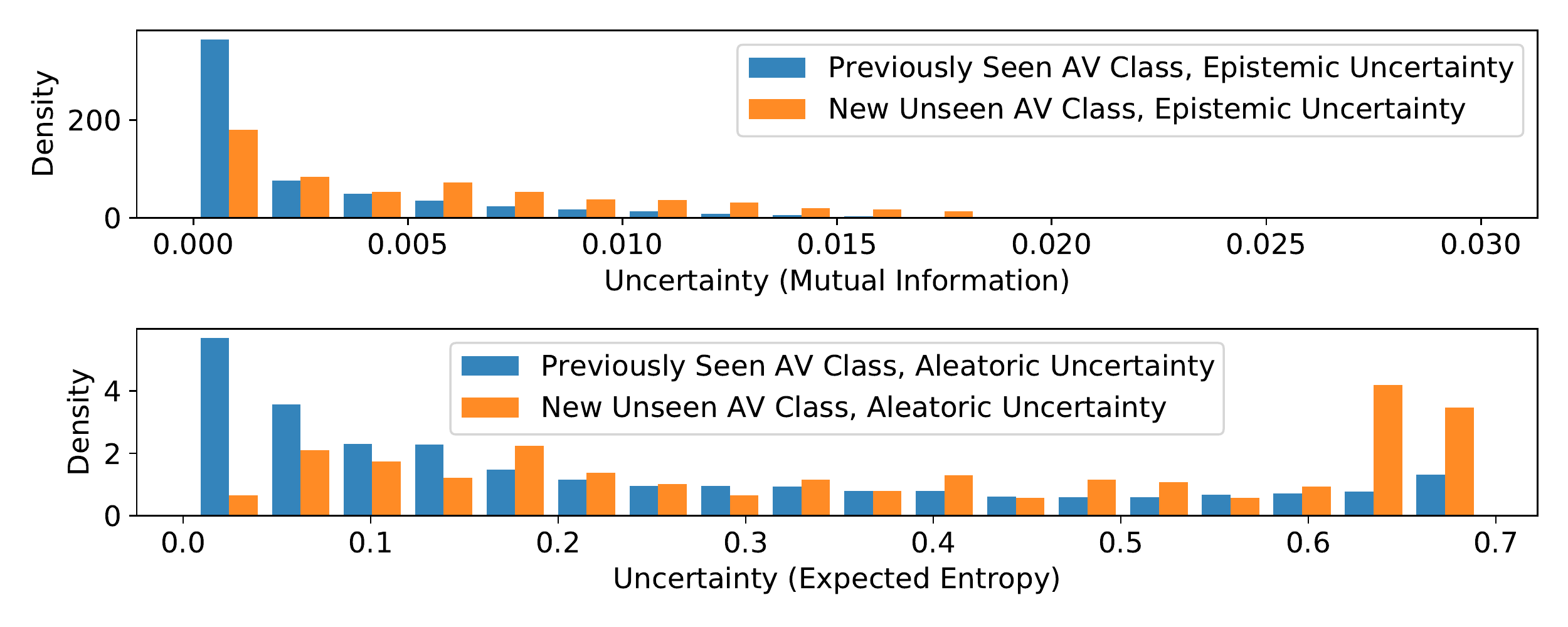}
        \caption{LGBM} \label{fig:figs/unseen_ember2018_v1_test_lgbm.pdf}
    \end{subfigure}
    \caption{A comparison of uncertainty distributions for the EMBER2018 models at test time between malware families seen and unseen during training.}
\end{figure}

Our local threshold adjustments using epistemic and aleatoric uncertainty estimates show improved TPR for extremely low target FPRs. We further investigate how this is possible, and in doing so show that these uncertainty estimates provide an additional benefit to practical application. We observe that the errors of our models are correlated with both uncertainty measures. This means we can use the uncertainty measures not just as a means of adjusting the threshold, but as a sourcing tool for analysts. The data with the highest uncertainty scores are the most likely to be mispredicted, and thus guide the analysts to the samples where their time is best spent. Additional details can be found in the Appendix, and next we focus on an extended version of this result, in that uncertainties are also correlated with a sample being novel malware or not.

Of the 200,000 files in the EMBER2018 test set, 363 belong to new malware families that were not present in the train set (we note that all the test set are new files that did not exist prior, as the train/test split is a split in time). \autoref{fig:figs/unseen_ember2018_v1_test_bmc.pdf} and \autoref{fig:figs/unseen_ember2018_v1_test_lgbm.pdf} show that the Bayesian MalConv and LightGBM ensemble uncertainty distributions for test samples from malware families seen during training place most of their weight on lower uncertainty values, whereas the uncertainty distributions for novel families not seen during training place most of their weight on higher uncertainty values. 
We however found that the Bayesian logistic regression model mostly does not exhibit this behavior, likely due to the simplicity of the model class which limits the extent to which predictions can disagree. 
Overall, these results suggest that for some models, uncertainties can be leveraged for the detection of new, out of training distribution, malware families.

\section{Conclusions} \label{sec:conclusion}

We have provided evidence that uncertainty estimation using ensembling and Bayesian methods can lead to significant improvements in machine learning based malware detection systems. In particular, improvements were especially large under extreme false positive rate constraints which are common in deployed, production scenarios. Local uncertainty based threshold adjustments were shown to lead to higher TPRs while satisfying desired FPR maximums. We additionally demonstrated how previous works have used an evaluation protocol that can lead to misleading results, and how uncertainty can be used to better detect model errors and new malware families.

Obtaining uncertainties has an inherent additional computational cost at prediction time which may limit use in resource limited deployed contexts. However, recent advances such as BatchEnsemble \cite{Wen2020Batchensemble:Learning} have introduced new methods to avoid the computational and memory costs of naive ensembles. 

We are currently working with professional malware analysts and teams that believe this approach may benefit them in production environments based on the evidence this work has provided. 
Future work includes leveraging uncertainty estimates to decide when to run more expensive malware analysis algorithms and techniques such as dynamic analysis, exploring and explaining malware specific drivers of uncertainty, and evaluating these methods over a long period of time in production.

\bibliographystyle{named}
\bibliography{Mendeley-Andre,RaffAdd}

\clearpage
\appendix

\section{Reproducibility}

The Sophos dataset and pre-trained models can be found here:
\url{https://github.com/sophos-ai/SOREL-20M}

The EMBER2018 dataset and a description of the EMBER2018 featurization method can be found here:
\url{https://github.com/elastic/ember}

\section{Additional Results and Figures}

Results comparing our new uncertainty augmented local adjustments \autoref{eq:lv1} (g+l) and \autoref{eq:lv2} (g+lv2) and \autoref{eq:lv3} (g+lv3) against the naive approach (g) are provided in \autoref{tab:ember-val} for the EMBER2018 dataset. 

The only case where (g) performed best is when using the Bayesian Logistic Regression (BLR) model on the EMBER2018 corpus at a target FPR of 0.01\%. In this one case we have pushed the model beyond what it is capable of achieving, and all three methods perform poorly - by happenstance the global threshold's degenerate solution of claiming that there is no malware receives a better score due to our uncertainty approaches failing to meet the FPR goal, which has a high penalty. However, we would argue our uncertainty based approaches are still preferable in this scenario because the degenerate model (g) is equivalent to having no anti-virus installed. 

\begin{table*}[!tb]
\centering
\caption{EMBER2018 comparison of the standard global adjustment (labeled as (g)) to the uncertainty aware local adjustments from \autoref{eq:lv1} (labeled as (g+l)) and \autoref{eq:lv2} (labeled as (g+lv2)) and \autoref{eq:lv3} (labeled as (g+lv3)). Best combined score (TPR penalized for over-runs on target FPR) shown in \textbf{bold}.}
\label{tab:ember-val}
\adjustbox{max width=2\columnwidth}{
\begin{tabular}{@{}lrrrrrrrrr@{}}
\toprule
             & \multicolumn{3}{c}{Target FPR=1\%}                           & \multicolumn{3}{c}{Target FPR=0.1\%}                         & \multicolumn{3}{c}{Target FPR=0.01\%}                        \\ \cmidrule(lr){2-4} \cmidrule(l){5-7}  \cmidrule(l){8-10} 
Method       & TPR                & FPR                & Comb.              & TPR                & FPR                & Comb.              & TPR                & FPR                & Comb.              \\ \midrule
BMC (g)      & 7.602E-01          & 1.177E-02          & 5.832E-01          & 4.998E-01          & 8.100E-04          & 4.998E-01          & 2.422E-01          & 8.000E-05          & 2.422E-01          \\
BMC (g+l)    & 7.617E-01          & 1.217E-02          & 5.447E-01          & 4.998E-01          & 8.300E-04          & 4.998E-01          & 2.431E-01          & 9.000E-05          & 2.431E-01          \\
BMC (g+lv2)  & \textbf{7.594E-01} & \textbf{1.166E-02} & \textbf{5.934E-01} & 5.016E-01 & 8.200E-04 & 5.016E-01 & \textbf{2.434E-01} & \textbf{9.000E-05} & \textbf{2.434E-01} \\ 
BMC (g+lv3) & 7.605E-01 & 1.173E-02 & 5.875E-01 & \textbf{5.023E-01} & \textbf{8.200E-04} & \textbf{5.023E-01} &  \textbf{2.434E-01} & \textbf{9.000E-05} & \textbf{2.434E-01} \\ \midrule
BLR (g)      & 7.778E-01          & 9.550E-03          & 7.778E-01          & 0.000E+00          & 0.000E+00          & 0.000E+00          & \textbf{0.000E+00} & \textbf{0.000E+00} & \textbf{0.000E+00} \\
BLR (g+l)    & \textbf{7.781E-01} & \textbf{9.560E-03} & \textbf{7.781E-01} & \textbf{5.977E-02} & \textbf{7.500E-04} & \textbf{5.977E-02} & 1.000E-04          & 8.600E-04          & -7.600E+00         \\
BLR (g+lv2)  & \textbf{7.781E-01} & \textbf{9.560E-03} & \textbf{7.781E-01} & 5.248E-02          & 7.800E-04          & 5.248E-02          & 9.000E-05          & 7.900E-04          & -6.900E+00         \\ 
BLR (g+lv3) & 7.779E-01 & 9.290E-03 & 7.779E-01 & 5.261E-02 & 8.200E-04 & 5.261E-02 &  5.840E-03 & 1.200E-04 & -1.942E-01 \\ \midrule
LGBM (g)     & 8.805E-01          & 1.805E-02          & 7.547E-02          & 6.954E-01          & 1.550E-03          & 1.454E-01          & 4.888E-01          & 8.000E-05          & 4.888E-01          \\
LGBM (g+l)   & 8.680E-01          & 1.494E-02          & 3.740E-01          & 6.892E-01 & 1.390E-03 & 2.992E-01 & \textbf{5.142E-01} & \textbf{9.000E-05} & \textbf{5.142E-01} \\
LGBM (g+lv2) & \textbf{8.693E-01} & \textbf{1.488E-02} & \textbf{3.813E-01} & 6.917E-01          & 1.430E-03          & 2.617E-01          & \textbf{5.142E-01} & \textbf{9.000E-05} & \textbf{5.142E-01} \\ 
LGBM (g+lv3) & 8.727E-01 & 1.512E-02 & 3.607E-01 & \textbf{6.890E-01} & \textbf{1.380E-03} & \textbf{3.090E-01} &  5.136E-01 & 9.000E-05 & 5.136E-01 \\ \bottomrule
\end{tabular}
}
\end{table*}

\begin{figure}[!tb]
    \centering
    \begin{subfigure}[t]{\columnwidth}
        \centering
        \includegraphics[width=\columnwidth]{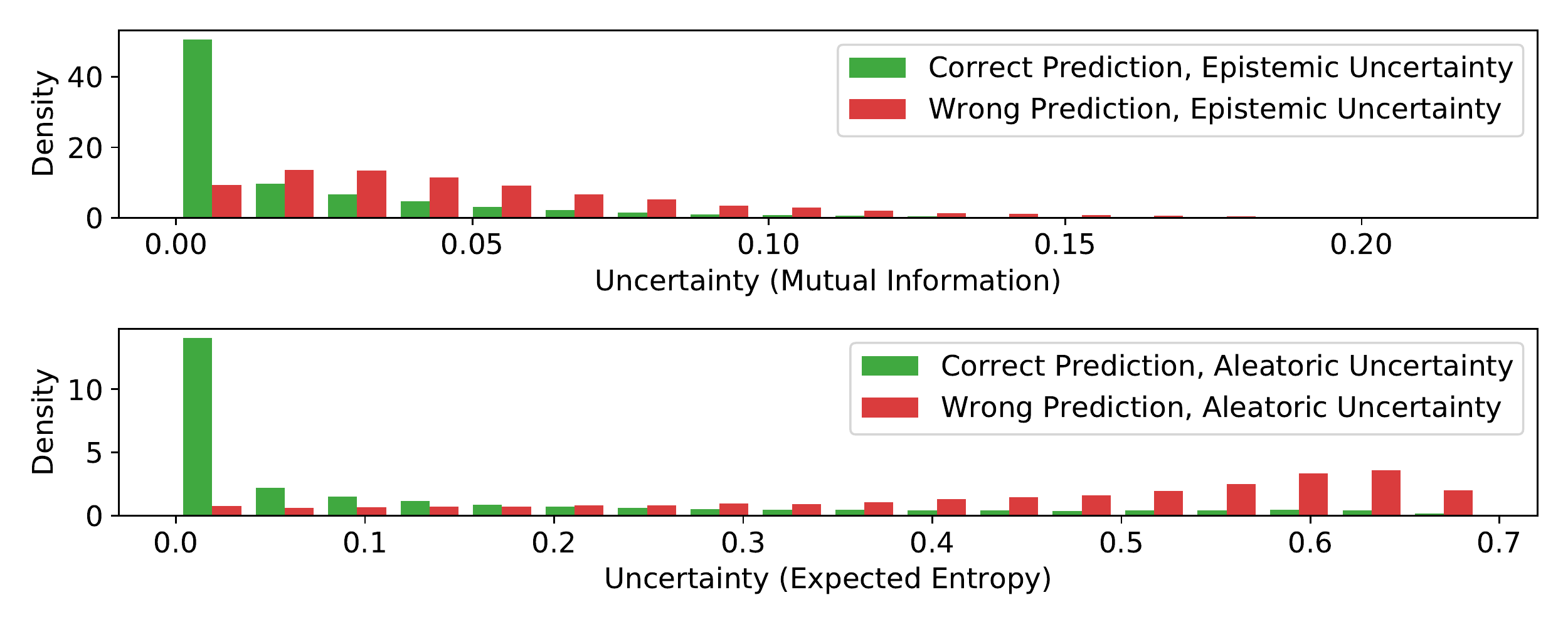}
        \caption{EMBER, Bayesian MalConv} \label{fig:figs/err_ember2018_v1_test_bmc.pdf}
    \end{subfigure}%
    \\
    \begin{subfigure}[t]{1\columnwidth}
        \centering
        \includegraphics[width=\columnwidth]{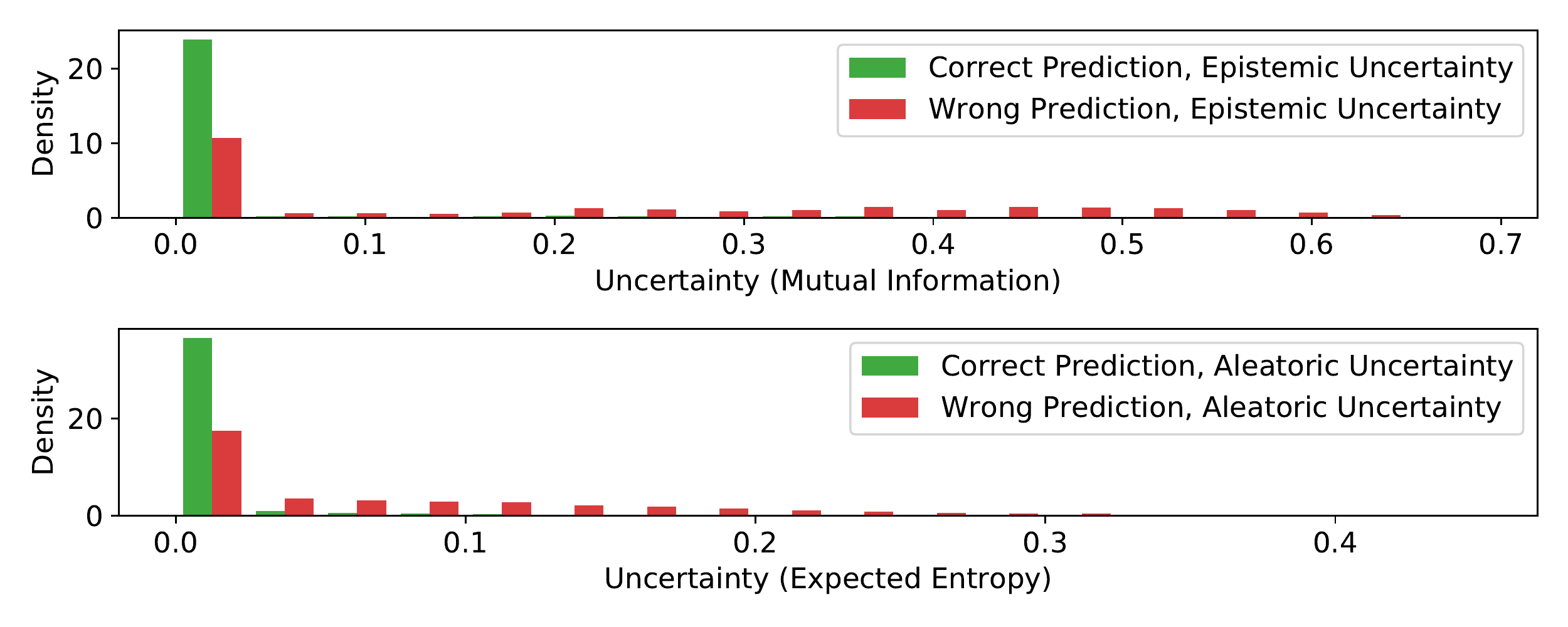}
        \caption{EMBER, Bayesian Logistic Regression} \label{fig:figs/err_ember2018_v1_test_csl.pdf}
    \end{subfigure}
    \\
    \begin{subfigure}[t]{1\columnwidth}
        \centering
        \includegraphics[width=\columnwidth]{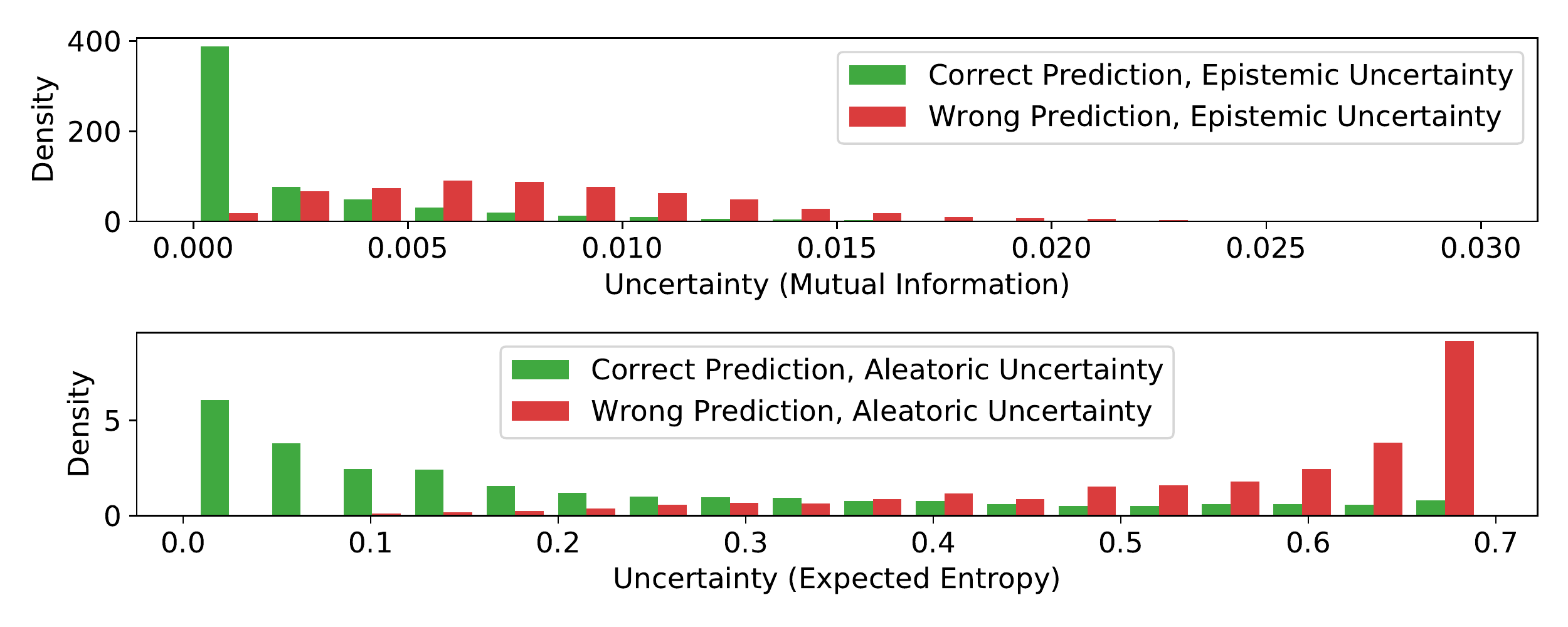}
        \caption{EMBER, LGBM Ensemble} \label{fig:figs/err_ember2018_v1_test_lgbm.pdf}
    \end{subfigure}
    \caption{A comparison of uncertainty distributions for all three EMBER2018 models at test time between samples predicted correctly and incorrectly.}
\end{figure}

\begin{figure}[!tb]
    \centering
    \begin{subfigure}[t]{\columnwidth}
        \centering
        \includegraphics[width=\columnwidth]{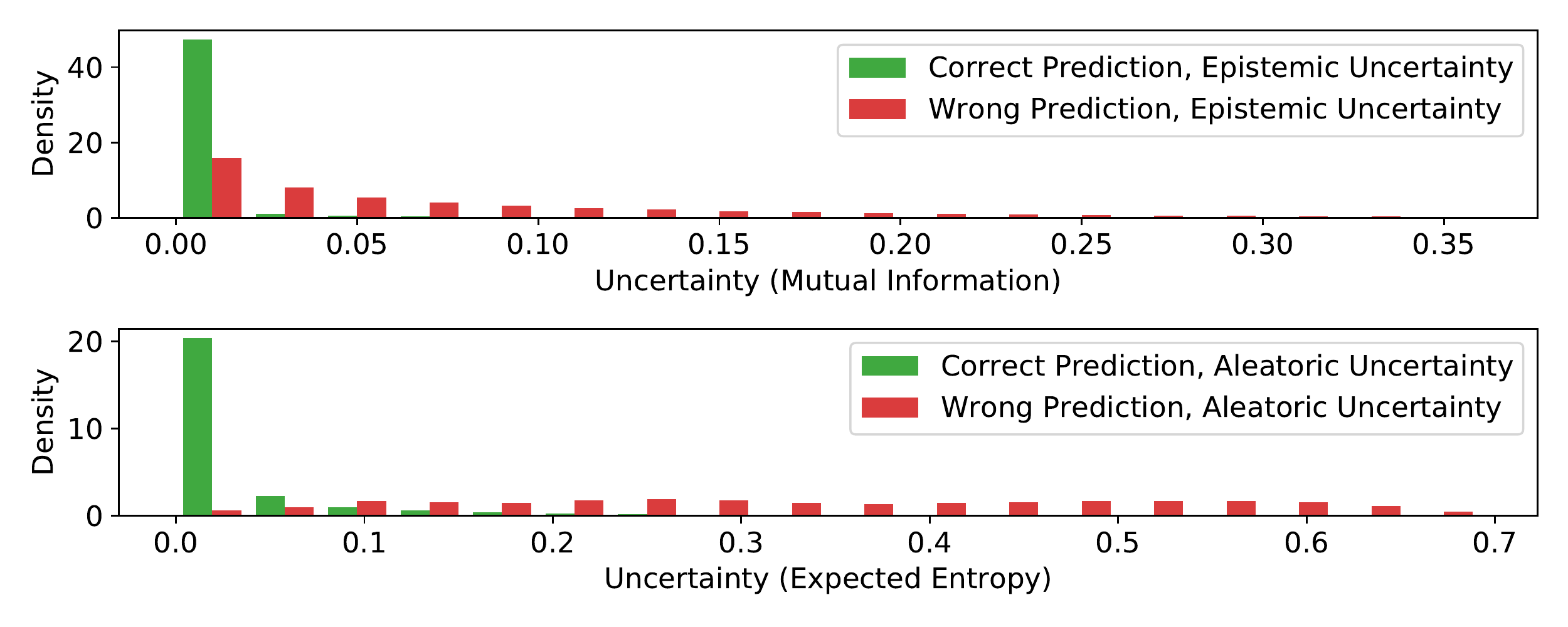}
        \caption{Sophos, FFNN Ensemble} \label{fig:figs/err_sophos_v1_test_ffnn.pdf}
    \end{subfigure}%
    \\
    \begin{subfigure}[t]{1\columnwidth}
        \centering
        \includegraphics[width=\columnwidth]{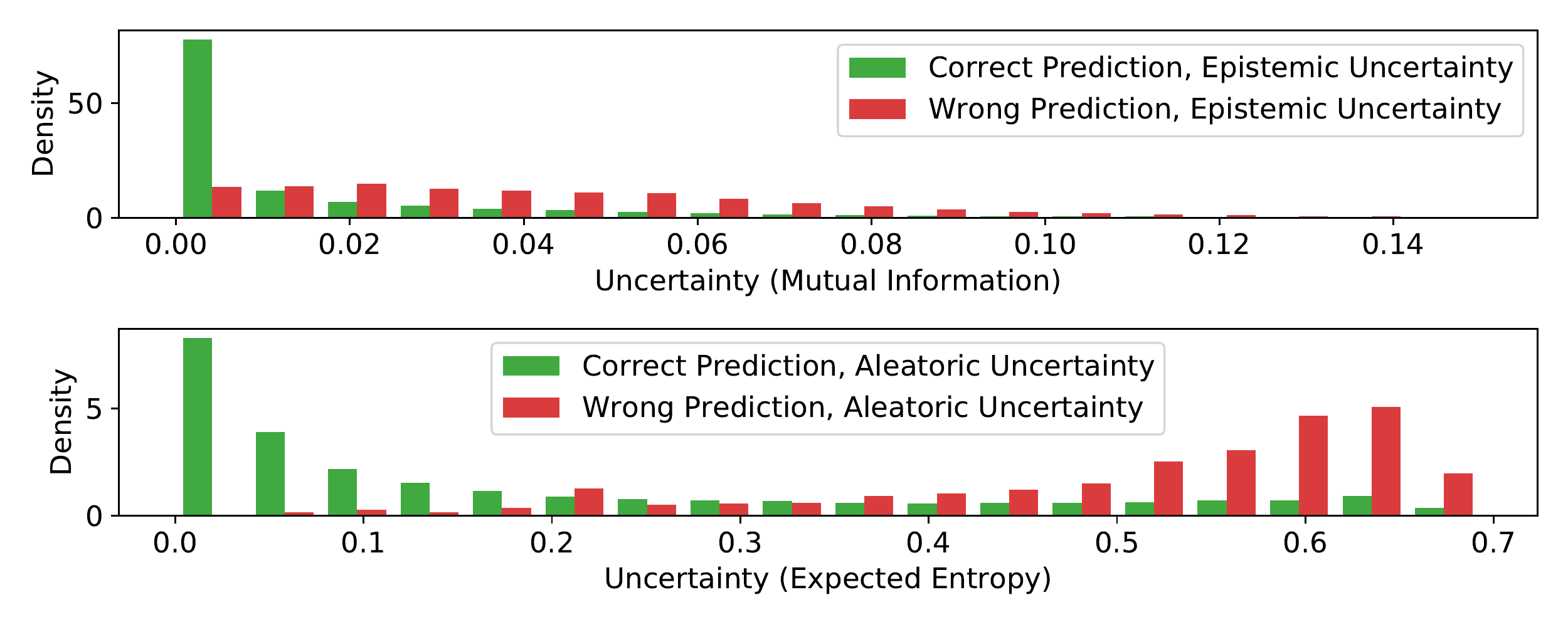}
        \caption{Sophos, LGBM Ensemble} \label{fig:figs/err_sophos_v1_test_lgbm.pdf}
    \end{subfigure}
    \caption{A comparison of uncertainty distributions for an ensemble of Sophos models at test time on Sophos data between samples predicted correctly and incorrectly.}

\end{figure}

\autoref{fig:figs/err_ember2018_v1_test_bmc.pdf}, \autoref{fig:figs/err_ember2018_v1_test_csl.pdf}, and  \autoref{fig:figs/err_ember2018_v1_test_lgbm.pdf} on the EMBER2018 dataset and \autoref{fig:figs/err_sophos_v1_test_ffnn.pdf} and \autoref{fig:figs/err_sophos_v1_test_lgbm.pdf} on the Sophos data show that the uncertainty distributions for test samples that the models ultimately got wrong place most of their weight on higher uncertainties. Consistently, the uncertainty distribution for test samples that a model ultimately got right places most of its weight on lower uncertainties. This suggests that overall system performance can be improved by leveraging uncertainty and flagging high uncertainty predictions for further processing and review. This explains the success of our approach, which can learn to use the uncertainty terms as a kind of additional offset. The more we want to lower the FPR rate, the less we should trust the model's outputs if uncertainty is high. 

\end{document}

%% file: figs/in_valid.tex
\begin{tikzpicture}

\definecolor{color0}{rgb}{0.698039215686274,0.133333333333333,0.133333333333333}
\definecolor{color1}{rgb}{0.419607843137255,0.556862745098039,0.137254901960784}

\begin{axis}[
legend cell align={left},
legend style={fill opacity=0.8, draw opacity=1, text opacity=1, draw=white!80!black,
legend columns=1, 
font=\small
},
log basis x={10},
tick align=outside,
tick pos=left,
x grid style={white!69.0196078431373!black},
xlabel={False Positive Rate (log scale)},
xmajorgrids,
xmin=5.31969245041414e-07, xmax=1.98960703710449,
xmode=log,
xtick style={color=black},
y grid style={white!69.0196078431373!black},
ylabel={\tiny $|$(Invalid TPR - Valid TPR) / Valid TPR$|$ },
ymajorgrids,
ymin=0, ymax=0.4,
ytick style={color=black},
width=\columnwidth,
height=\columnwidth,
]
\addplot [very thick, color0]
table {%
0 0.0277275126291184
0 0.0277275126291184
0 0.0277275126291184
0 0.0277275126291184
0 0.351127195958682
0 0.351127195958682
2.11681950691512e-06 0.174643573053447
2.11681950691512e-06 0.174643573053447
2.11681950691512e-06 0.269886500845207
2.11681950691512e-06 0.192253078966433
2.11681950691512e-06 0.0857106341801849
4.93924551613528e-06 0.0524447400578508
5.2920487672878e-06 0.0111634792591922
5.99765526959284e-06 0.0480580072810595
7.76167152535544e-06 0.0883524741244861
9.87849103227056e-06 0.125752236200526
1.41121300461008e-05 0.172841737102885
2.04625885668462e-05 0.138465378854795
2.78714568410491e-05 0.131056146117202
4.12779803848448e-05 0.113900697555517
5.32732909240305e-05 0.113260225625512
7.47942892443342e-05 0.0832696864657383
9.80793038204006e-05 0.0814536211100698
0.000128420383419517 0.0691060576883232
0.000186632919859683 0.0584029724925321
0.000307291631753845 0.0461416568935494
0.000426186327392244 0.0519935919565866
0.0006159944765123 0.0517340711490137
0.000905645945708519 0.0236369130883136
0.00130501922601317 0.0112352519540807
0.00246785874181188 0.0321888580553957
0.00438393319882121 0.072668114099993
0.00622944700560005 0.101186739637855
0.0083646122815751 0.11646546208366
0.0108416439079169 0.0541589515015168
0.0144751645915367 0.0161214146521513
0.0190143312208651 0.0588249064327977
0.024431625142312 0.0386032196842353
0.0311945106636547 0.0192457581909005
0.0415993841466448 0.0173839098110606
0.0556074372336556 0.00798215650640506
0.0753150268430354 0.00293708814435896
0.10222685884095 0.000762017669619791
0.137728391594674 0.000273523182600924
0.185375881875827 0.000848602053571977
0.247595910445834 0.00121580049257406
0.335454151277095 0.00152243797715223
0.454199258337005 0.00222957289536516
0.634885679398513 0.000423555884007465
0.999999294393498 0
};
\addlegendentry{Sophos, LGBM}
\addplot [very thick, color1]
table {%
0 2.48591817610775
0 2.48591817610775
0 2.48591817610775
0 2.48591817610775
0 2.61568205455613
0 2.61568205455613
0 0.195820962920777
0 0.195820962920777
1.05840975345756e-06 0.0430341286381895
1.05840975345756e-06 0.0870837717787703
3.88083576267772e-06 0.0465190937186311
3.88083576267772e-06 0.00232588174654056
6.70326177189788e-06 0.0205775939677202
8.46727802766048e-06 0.0118916371686266
1.02312942834231e-05 0.0111017010006632
1.6934556055321e-05 0.0251186734572451
2.01097853156936e-05 0.0142785073452358
2.61074405852865e-05 0.00344792493154097
4.05723738825398e-05 0.00519220967229786
5.96237494447759e-05 0.0078920790008765
8.71424030346724e-05 0.012464046856245
0.000116425072880332 0.00828051837892004
0.000161231085776702 0.00790884301199024
0.000226146883988765 0.00654453520935042
0.000306233222000387 0.00595374136285629
0.000429714359903769 0.00411253168778107
0.00061458326350769 0.00536721130545365
0.000834732492226862 0.00420396566431054
0.0011695427775706 0.00343876195129082
0.0015745609098937 0.00252475084901406
0.00215456945478844 0.00214321862151473
0.0029335590333332 0.00218973480736356
0.00390694320326301 0.0015255244037978
0.00523665865685685 0.000239354790667162
0.0071534387203685 7.64010130319663e-05
0.00969750296442932 0.000284233771344869
0.0132763391441205 0.00035936066880182
0.0182152318570046 0.00045663264541006
0.0249093207443725 0.000651369333890776
0.0342201513455387 0.000744267906616087
0.0474051144476107 0.000713927688049237
0.0654143092059422 0.000585342837302859
0.0907533443102185 0.00051015799348712
0.125950407858198 0.000580916236461017
0.175503388498826 0.000478030866925189
0.24453675345509 0.000344743762568721
0.339179753609265 0.000229584628426068
0.475117668704341 0.000127945598119963
0.678088907124897 2.13168835597805e-05
0.999998941590246 0
};
\addlegendentry{Sophos, FFNN}
\addplot [very thick, blue!54.5098039215686!black]
table {%
0 0.0213779781068899
0 0.0213779781068899
0 0.0213779781068899
0 0.0213779781068899
0 0.0213779781068899
0 0.0213779781068899
0 0.0213779781068899
0 0.0213779781068899
0 0.0213779781068899
0 0.0213779781068899
0 0.0213779781068899
1e-05 0.278759586528843
1e-05 0.165275049115914
1e-05 0.157531336307627
1e-05 0.369657769772392
1e-05 0.369657769772392
1e-05 0.34283191523519
4e-05 0.014351599873958
4e-05 0.0402032250938811
0.0001 0.0627590591226954
0.00016 0.0475922482137613
0.00027 0.126150586053904
0.00032 0.172492742957603
0.00049 0.171199726687798
0.00071 0.211556920714149
0.00096 0.193027169207143
0.00145 0.155700858193707
0.00182 0.132878968763695
0.00229 0.105337303028103
0.00311 0.117001521857504
0.00401 0.103300939467051
0.00553 0.0938883740088559
0.00754 0.087459855104937
0.01073 0.0750907386486551
0.01466 0.0684208683064638
0.01911 0.05787810692735
0.02557 0.0532900370316826
0.03463 0.0437734286461512
0.04626 0.0322971769477558
0.06424 0.0276672751638194
0.08829 0.0192865641111274
0.12193 0.0162472981498144
0.16653 0.0133047474457347
0.22613 0.0138774514289199
0.30193 0.0106784934938981
0.41002 0.00867644105242014
0.53705 0.00511788153756791
0.68856 0.00116069641785065
0.85245 0.000340013600544029
1 0
};
\addlegendentry{EMBER2018, LGBM}
\end{axis}

\end{tikzpicture}

%% file: figs/subsample.tex
\begin{tikzpicture}

\definecolor{color0}{rgb}{0.12156862745098,0.466666666666667,0.705882352941177}
\definecolor{color1}{rgb}{1,0.498039215686275,0.0549019607843137}
\definecolor{color2}{rgb}{0.172549019607843,0.627450980392157,0.172549019607843}
\definecolor{color3}{rgb}{0.83921568627451,0.152941176470588,0.156862745098039}
\definecolor{color4}{rgb}{0.580392156862745,0.403921568627451,0.741176470588235}
\definecolor{color5}{rgb}{0.549019607843137,0.337254901960784,0.294117647058824}
\definecolor{color6}{rgb}{0.890196078431372,0.466666666666667,0.76078431372549}
\definecolor{color7}{rgb}{0.737254901960784,0.741176470588235,0.133333333333333}
\definecolor{color8}{rgb}{0.0901960784313725,0.745098039215686,0.811764705882353}

\begin{axis}[
legend cell align={left},
legend style={
  fill opacity=0.8,
  draw opacity=1,
  text opacity=1,
  at={(0.03,0.03)},
  anchor=south west,
  draw=white!80!black,
  legend columns=2, 
  font=\tiny
},
log basis x={10},
log basis y={10},
tick align=outside,
tick pos=left,
x grid style={white!69.0196078431373!black},
xlabel={False Positive Rate},
xmajorgrids,
xmin=5.31969263809468e-07, xmax=1.98960556303087,
xmode=log,
xtick style={color=black},
y grid style={white!69.0196078431373!black},
ylabel={$|$(Invalid TPR - Valid TPR) / Valid TPR$|$},
ymajorgrids,
ymin=5.52448833704581e-06, ymax=4.87345539289546,
ymode=log,
ytick style={color=black}
]
\addplot [very thick, color0]
table {%
0 0.0277275126291184
0 0.0277275126291184
0 0.0277275126291184
0 0.0277275126291184
0 0.351127195958682
0 0.351127195958682
2.11681950691512e-06 0.174643573053447
2.11681950691512e-06 0.174643573053447
2.11681950691512e-06 0.269886500845207
2.11681950691512e-06 0.192253078966433
2.11681950691512e-06 0.0857106341801849
4.93924551613528e-06 0.0524447400578508
5.2920487672878e-06 0.0111634792591922
5.99765526959284e-06 0.0480580072810595
7.76167152535544e-06 0.0883524741244861
9.87849103227056e-06 0.125752236200526
1.41121300461008e-05 0.172841737102885
2.04625885668462e-05 0.138465378854795
2.78714568410491e-05 0.131056146117202
4.12779803848448e-05 0.113900697555517
5.32732909240305e-05 0.113260225625512
7.47942892443342e-05 0.0832696864657383
9.80793038204006e-05 0.0814536211100698
0.000128420383419517 0.0691060576883232
0.000186632919859683 0.0584029724925321
0.000307291631753845 0.0461416568935494
0.000426186327392244 0.0519935919565866
0.0006159944765123 0.0517340711490137
0.000905645945708519 0.0236369130883136
0.00130501922601317 0.0112352519540807
0.00246785874181188 0.0321888580553957
0.00438393319882121 0.072668114099993
0.00622944700560005 0.101186739637855
0.0083646122815751 0.11646546208366
0.0108416439079169 0.0541589515015168
0.0144751645915367 0.0161214146521513
0.0190143312208651 0.0588249064327977
0.024431625142312 0.0386032196842353
0.0311945106636547 0.0192457581909005
0.0415993841466448 0.0173839098110606
0.0556074372336556 0.00798215650640506
0.0753150268430354 0.00293708814435896
0.10222685884095 0.000762017669619791
0.137728391594674 0.000273523182600924
0.185375881875827 0.000848602053571977
0.247595910445834 0.00121580049257406
0.335454151277095 0.00152243797715223
0.454199258337005 0.00222957289536516
0.634885679398513 0.000423555884007465
0.999999294393498 0
};
\addlegendentry{LGBM, 100.0\% }
\addplot [very thick, color1]
table {%
0 2.48591817610775
0 2.48591817610775
0 2.48591817610775
0 2.48591817610775
0 2.61568205455613
0 2.61568205455613
0 0.195820962920777
0 0.195820962920777
1.05840975345756e-06 0.0430341286381895
1.05840975345756e-06 0.0870837717787703
3.88083576267772e-06 0.0465190937186311
3.88083576267772e-06 0.00232588174654056
6.70326177189788e-06 0.0205775939677202
8.46727802766048e-06 0.0118916371686266
1.02312942834231e-05 0.0111017010006632
1.6934556055321e-05 0.0251186734572451
2.01097853156936e-05 0.0142785073452358
2.61074405852865e-05 0.00344792493154097
4.05723738825398e-05 0.00519220967229786
5.96237494447759e-05 0.0078920790008765
8.71424030346724e-05 0.012464046856245
0.000116425072880332 0.00828051837892004
0.000161231085776702 0.00790884301199024
0.000226146883988765 0.00654453520935042
0.000306233222000387 0.00595374136285629
0.000429714359903769 0.00411253168778107
0.00061458326350769 0.00536721130545365
0.000834732492226862 0.00420396566431054
0.0011695427775706 0.00343876195129082
0.0015745609098937 0.00252475084901406
0.00215456945478844 0.00214321862151473
0.0029335590333332 0.00218973480736356
0.00390694320326301 0.0015255244037978
0.00523665865685685 0.000239354790667162
0.0071534387203685 7.64010130319663e-05
0.00969750296442932 0.000284233771344869
0.0132763391441205 0.00035936066880182
0.0182152318570046 0.00045663264541006
0.0249093207443725 0.000651369333890776
0.0342201513455387 0.000744267906616087
0.0474051144476107 0.000713927688049237
0.0654143092059422 0.000585342837302859
0.0907533443102185 0.00051015799348712
0.125950407858198 0.000580916236461017
0.175503388498826 0.000478030866925189
0.24453675345509 0.000344743762568721
0.339179753609265 0.000229584628426068
0.475117668704341 0.000127945598119963
0.678088907124897 2.13168835597805e-05
0.999998941590246 0
};
\addlegendentry{FFNN, 100.0\% }
\addplot [very thick, color2]
table {%
9.87849103227056e-06 0.642188228036672
9.87849103227056e-06 0.642188228036672
9.87849103227056e-06 0.642188228036672
9.87849103227056e-06 0.642188228036672
9.87849103227056e-06 0.529593972922778
9.87849103227056e-06 0.529593972922778
9.87849103227056e-06 0.519028212548973
9.87849103227056e-06 0.519028212548973
9.87849103227056e-06 0.503967082078239
9.87849103227056e-06 0.451223593802296
9.87849103227056e-06 0.219498749827732
9.87849103227056e-06 0.195971885889788
9.87849103227056e-06 0.140839617007593
1.30537202926432e-05 0.136738393533407
1.30537202926432e-05 0.0468790742143362
1.97569820645411e-05 0.152239233787671
1.97569820645411e-05 0.00565862433653253
3.06938828502692e-05 0.0543906027584062
3.31635056083369e-05 0.0507732327351296
5.08036681659629e-05 0.00808609637953856
6.45629949609112e-05 0.0244007109987329
8.14975510162321e-05 0.0561394129664533
9.45512713088754e-05 0.0945576680735566
0.000110780220861891 0.10934365981298
0.000228616506746833 0.02458620139086
0.000360212119426723 0.00789112741612901
0.000478048405311665 0.0224909939884518
0.000645982752860264 0.0315795620839144
0.000953627187865262 0.0133575131770896
0.0013515892551653 0.00591344303854791
0.00269083039654027 0.0432967146720495
0.00457303574143896 0.0791138825621462
0.00635575056951265 0.108252074285006
0.00847574530568814 0.119314663180348
0.0108748074135253 0.0549037845352148
0.0144751645915367 0.0161214146521513
0.018833695956275 0.0569528507856326
0.0243423659197704 0.0380107359046399
0.0309863567454747 0.0182654656133321
0.0417429750698639 0.0180041254125413
0.0557238623065359 0.00813630107881506
0.0750564220599406 0.00277432641737435
0.10222685884095 0.000762017669619791
0.13781588680096 0.000257651727399958
0.184943345089914 0.000907076016865222
0.247059296700831 0.0012448729030927
0.334934472088147 0.00154100573242269
0.453194474677723 0.00225547917485203
0.633909119999323 0.000443418869789304
0.99999400234473 0
};
\addlegendentry{LGBM, 10.0\% }
\addplot [very thick, color3]
table {%
6.70326177189788e-06 0.244566684891155
6.70326177189788e-06 0.244566684891155
6.70326177189788e-06 0.244566684891155
6.70326177189788e-06 0.244566684891155
6.70326177189788e-06 0.216445555270411
6.70326177189788e-06 0.216445555270411
6.70326177189788e-06 0.208124030661772
6.70326177189788e-06 0.208124030661772
6.70326177189788e-06 0.14506767236089
6.70326177189788e-06 0.108961985204691
6.70326177189788e-06 0.0853475112415339
6.70326177189788e-06 0.0432589848587973
6.70326177189788e-06 0.037373218928177
1.27009170414907e-05 0.0652066855151665
1.27009170414907e-05 0.0291063136253503
1.97569820645411e-05 0.0390342377579365
1.97569820645411e-05 0.00233318284780383
2.5754637334134e-05 0.00568537583450841
5.43317006774881e-05 0.0271169786060364
7.23246664862666e-05 0.0182919634759484
8.96120257927401e-05 0.0143794452177043
0.000123128334652229 0.0109571682086066
0.000156291840260566 0.00657652519189955
0.000262485618857475 0.0127393841215063
0.000393022821783907 0.0154344827586208
0.000515092746682679 0.0105474187532719
0.000743709253429512 0.0109021471058171
0.000945512713088754 0.00713432625828853
0.00131419211054314 0.00572518484796157
0.00162148374229698 0.00317888103700152
0.00224277026757657 0.00291677444361843
0.00302317105912594 0.00296572443451408
0.0040815808125835 0.00281928950540884
0.00544834060754837 0.000980247742241129
0.00724798999167737 8.31957452183271e-05
0.00976100754963677 0.000229183849172914
0.0132548181458002 0.000375154403478113
0.0178994729472231 0.000585675033595616
0.0247099869074714 0.000732195277003919
0.0344000810036265 0.000704714803037873
0.047306329537288 0.000731788941360211
0.0659230514941041 0.000527445723535342
0.0905402511465224 0.000519788619294838
0.125056404419778 0.000626028290270161
0.176395980724242 0.000453675658050957
0.244725503194457 0.000343269992898773
0.341164977503501 0.00021780849335385
0.476279097007135 0.0001250039523309
0.679203412595288 2.05818034804579e-05
0.9999848294602 0
};
\addlegendentry{FFNN, 10.0\% }
\addplot [very thick, color4]
table {%
5.11564714171154e-05 0.824696161018584
5.11564714171154e-05 0.824696161018584
5.11564714171154e-05 0.824696161018584
5.11564714171154e-05 0.824696161018584
5.11564714171154e-05 0.769532505948171
5.11564714171154e-05 0.769532505948171
5.11564714171154e-05 0.764355989968491
5.11564714171154e-05 0.764355989968491
5.11564714171154e-05 0.75697704327696
5.11564714171154e-05 0.731136261333676
5.11564714171154e-05 0.617606584785544
5.11564714171154e-05 0.606079994855636
5.11564714171154e-05 0.579068870169121
5.11564714171154e-05 0.55292585685808
5.11564714171154e-05 0.506388656678027
5.11564714171154e-05 0.448456690888046
5.11564714171154e-05 0.353093048678542
5.11564714171154e-05 0.237489550511221
5.11564714171154e-05 0.125924377853514
5.11564714171154e-05 0.00169764002314959
0.000248373488811374 0.229422593704195
0.000248373488811374 0.155429870175194
0.000302705189488862 0.123370423915347
0.000302705189488862 0.0797178234130191
0.000498158190627358 0.140097076898453
0.000525324040966102 0.080993080245644
0.000590592642429319 0.0314245510778418
0.0006159944765123 0.0516585652849958
0.000983262660962073 0.0053054455778774
0.00150964511168163 0.0108012153454051
0.00273104996717166 0.0457218305769018
0.00486762645615132 0.0880515170185093
0.0064044374181717 0.110351330685345
0.00849409107474807 0.119852489874844
0.0109164381971613 0.0560964892921202
0.0149242831302539 0.0226544935929721
0.0199312668706105 0.070290994524197
0.0248832133037872 0.0420273387181781
0.0315310849652542 0.0210879537786906
0.0425371351882082 0.0212464484401919
0.0555114747493421 0.0078004245981425
0.0748810788441178 0.00266167344980727
0.102596596648157 0.000895920803663887
0.137966533789202 0.00022893323158522
0.184130839202509 0.00107428927598597
0.242912094483533 0.00153345219635928
0.33168127330927 0.00165985567117351
0.45328549791652 0.00225399877994565
0.632322916582141 0.000461075519316618
0.999667306534163 0
};
\addlegendentry{LGBM, 1.0\% }
\addplot [very thick, color5]
table {%
9.80793038204006e-05 0.415725081616503
9.80793038204006e-05 0.415725081616503
9.80793038204006e-05 0.415725081616503
9.80793038204006e-05 0.415725081616503
9.80793038204006e-05 0.393975351514058
9.80793038204006e-05 0.393975351514058
9.80793038204006e-05 0.387539233309716
9.80793038204006e-05 0.387539233309716
9.80793038204006e-05 0.338769543301406
9.80793038204006e-05 0.310844315495787
9.80793038204006e-05 0.292580168851017
9.80793038204006e-05 0.260027632676978
9.80793038204006e-05 0.255475403724457
9.80793038204006e-05 0.221901960090544
9.80793038204006e-05 0.191852934094944
9.80793038204006e-05 0.158684228921454
9.80793038204006e-05 0.126552827861885
9.80793038204006e-05 0.0810305383366716
9.80793038204006e-05 0.0618378232837666
9.80793038204006e-05 0.0378477924494271
0.000349628021892147 0.0759667720255144
0.000349628021892147 0.0546980319793174
0.000383849937253942 0.0423748497547347
0.000383849937253942 0.0279236916479875
0.000449118538717158 0.019728588519742
0.000777578365540154 0.0226797589128839
0.00117342361333328 0.021656120101826
0.00140027610382435 0.0151901215209721
0.00147118955730601 0.0079273335588887
0.00175978261674877 0.00474638597279846
0.00210164896711556 0.00165816584458344
0.00293779267234703 0.00223287702962496
0.00389530069597497 0.00143303873874216
0.005075074767829 0.000419717283912187
0.00715061629435928 8.39661049258846e-05
0.00967668757261132 0.000300071172660055
0.013379357693457 0.00030220655893477
0.0175865364634508 0.000735015619081825
0.0248137110633102 0.00069327745232878
0.0361595108171241 0.000255670594900067
0.0492516866641429 0.000400715191120331
0.0651740501919073 0.000604643364280165
0.0897789017305352 0.000576835358304342
0.127470284264164 0.000509189051770339
0.184132956022016 0.000233056738252877
0.251761458432192 0.000253378253958638
0.383244174071713 5.8115284540536e-05
0.522882289664876 1.02930504998754e-05
0.745561823301314 1.83758943548646e-05
0.999921677678244 0
};
\addlegendentry{FFNN, 1.0\% }
\addplot [very thick, color6]
table {%
0.00100443085603122 0.917147363814294
0.00100443085603122 0.917147363814294
0.00100443085603122 0.917147363814294
0.00100443085603122 0.917147363814294
0.00100443085603122 0.891075748550311
0.00100443085603122 0.891075748550311
0.00100443085603122 0.888629207746265
0.00100443085603122 0.888629207746265
0.00100443085603122 0.885141747407578
0.00100443085603122 0.872928798064649
0.00100443085603122 0.81927205533741
0.00100443085603122 0.813824323174364
0.00100443085603122 0.801058242867042
0.00100443085603122 0.78870245201133
0.00100443085603122 0.766707898223903
0.00100443085603122 0.739327915486451
0.00100443085603122 0.694256859431795
0.00100443085603122 0.639619980792375
0.00100443085603122 0.586891707898224
0.00100443085603122 0.526574599739846
0.00100443085603122 0.463156007245408
0.00100443085603122 0.411606936626995
0.00100443085603122 0.365137857255741
0.00100443085603122 0.333524295213898
0.00100443085603122 0.297902053270767
0.00100443085603122 0.239558893251802
0.00100443085603122 0.175758883526423
0.00158585061393058 0.170780333373372
0.00158585061393058 0.0927207635583941
0.00486762645615132 0.16076642895755
0.00486762645615132 0.127334896028818
0.00886594570146283 0.212307002207146
0.00945265750812947 0.175044454430278
0.0109926436994102 0.140252289898779
0.0144751645915367 0.0664135210328501
0.0180991595873754 0.0590680104340927
0.022856358625916 0.0884433759065668
0.0270402523813337 0.0538575216046795
0.0312234405302492 0.0194088084732026
0.0415626926085249 0.0172605841218149
0.052774779930152 0.00187160890553964
0.070852065715956 0.000803605744034195
0.103055240874656 0.00100913659100183
0.138095306975873 0.000198705017490605
0.190199055122333 0.00028894724392235
0.245637852401937 0.00134626682740452
0.340829461611655 0.00131675331338426
0.468043610715481 0.001866285928588
0.849935842728778 6.61482116831194e-05
0.999201606242642 0
};
\addlegendentry{LGBM, 0.1\% }
\addplot [very thick, white!49.8039215686275!black]
table {%
9.80793038204006e-05 0.415725081616503
9.80793038204006e-05 0.415725081616503
9.80793038204006e-05 0.415725081616503
9.80793038204006e-05 0.415725081616503
9.80793038204006e-05 0.393975351514058
9.80793038204006e-05 0.393975351514058
9.80793038204006e-05 0.387539233309716
9.80793038204006e-05 0.387539233309716
9.80793038204006e-05 0.338769543301406
9.80793038204006e-05 0.310844315495787
9.80793038204006e-05 0.292580168851017
9.80793038204006e-05 0.260027632676978
9.80793038204006e-05 0.255475403724457
9.80793038204006e-05 0.221901960090544
9.80793038204006e-05 0.191852934094944
9.80793038204006e-05 0.158684228921454
9.80793038204006e-05 0.126552827861885
9.80793038204006e-05 0.0810305383366716
9.80793038204006e-05 0.0618378232837666
9.80793038204006e-05 0.0378477924494271
9.80793038204006e-05 0.0216263237061962
9.80793038204006e-05 0.000893186154665297
9.80793038204006e-05 0.0175789391019639
9.80793038204006e-05 0.0329348371078718
9.80793038204006e-05 0.047208555946678
9.80793038204006e-05 0.0618222896115115
9.80793038204006e-05 0.0732550723912276
0.000777578365540154 0.00225502873791461
0.000777578365540154 0.00616538677833525
0.00139815928431744 0.000118388195677756
0.00139815928431744 0.00639845076109493
0.00184621941328114 0.00755099160853663
0.00198063745197025 0.0157180377447817
0.00776484675461581 0.00570780396531028
0.00776484675461581 0.000909107397096287
0.0146635615276522 0.00334958672905609
0.0218229979032903 0.00371822363849822
0.0218229979032903 0.000972608237295474
0.0367042390369036 0.00274866554076713
0.0367042390369036 0.000127446260160376
0.0508050793789675 0.000144934107008857
0.102064569345419 0.00200011839398565
0.102064569345419 0.000125053647274752
0.274579361503732 0.00314705525018498
0.274579361503732 0.00138870992088187
0.274579361503732 7.14235749340667e-05
0.911678175696725 0.000964280464500927
0.911678175696725 0.000347640746729317
0.911678175696725 0.000108040570336627
0.999921677678244 0
};
\addlegendentry{FFNN, 0.1\% }
\addplot [very thick, color7]
table {%
0.023649813137758 0.95087358573358
0.023649813137758 0.95087358573358
0.023649813137758 0.95087358573358
0.023649813137758 0.95087358573358
0.023649813137758 0.935414753872367
0.023649813137758 0.935414753872367
0.023649813137758 0.933964108695672
0.023649813137758 0.933964108695672
0.023649813137758 0.931896263552495
0.023649813137758 0.924654750953056
0.023649813137758 0.892839669469145
0.023649813137758 0.889609506140914
0.023649813137758 0.882040021588483
0.023649813137758 0.87471381293525
0.023649813137758 0.861672422694578
0.023649813137758 0.84543781102148
0.023649813137758 0.818713502983734
0.023649813137758 0.786317262407296
0.023649813137758 0.75505270527305
0.023649813137758 0.719288445025503
0.023649813137758 0.681685199184395
0.023649813137758 0.651119835004879
0.023649813137758 0.623566586866165
0.023649813137758 0.60482172832209
0.023649813137758 0.583699974140673
0.023649813137758 0.54910614127482
0.023649813137758 0.511276738949643
0.023649813137758 0.458848079607082
0.023649813137758 0.407906106315172
0.023649813137758 0.359333820485813
0.023649813137758 0.333812376616321
0.023649813137758 0.303823666420087
0.023649813137758 0.267221455853083
0.023649813137758 0.227344030865404
0.0242643964012657 0.155425322600335
0.0242643964012657 0.10992826316432
0.0412924453181421 0.169203925160104
0.0412924453181421 0.113606306970531
0.0412924453181421 0.0639450305661263
0.0412924453181421 0.01624765670719
0.0524149206139764 0.00130751405497786
0.0863514181455885 0.00723138029327557
0.0863514181455885 0.0049138573086696
0.27267775198002 0.0157821838563455
0.27267775198002 0.00653576306204701
0.27267775198002 0.000208168997539946
0.998997333160225 0.00740396671522292
0.998997333160225 0.00259587159402097
0.998997333160225 9.62794957012347e-05
0.998997333160225 0
};
\addlegendentry{LGBM, 0.01\% }
\addplot [very thick, color8]
table {%
0.993366240468579 0.502402577644636
0.993366240468579 0.502402577644636
0.993366240468579 0.502402577644636
0.993366240468579 0.502402577644636
0.993366240468579 0.483879431612895
0.993366240468579 0.483879431612895
0.993366240468579 0.478398114979767
0.993366240468579 0.478398114979767
0.993366240468579 0.436863434517449
0.993366240468579 0.41308092916328
0.993366240468579 0.397526278422663
0.993366240468579 0.369802928366585
0.993366240468579 0.36592602500915
0.993366240468579 0.337333219659832
0.993366240468579 0.311741982710848
0.993366240468579 0.2834938726553
0.993366240468579 0.25612918209466
0.993366240468579 0.217360148520309
0.993366240468579 0.201014682990573
0.993366240468579 0.180583586036386
0.993366240468579 0.166768580840233
0.993366240468579 0.147589852288145
0.993366240468579 0.133379439697432
0.993366240468579 0.120301597357679
0.993366240468579 0.108145392327921
0.993366240468579 0.095699613853076
0.993366240468579 0.08596289050155
0.993366240468579 0.0768009042923016
0.993366240468579 0.0690096147203264
0.993366240468579 0.0626977955670275
0.993366240468579 0.0565888248168852
0.993366240468579 0.0502152691930602
0.993366240468579 0.0412892044961789
0.993366240468579 0.0329988784541188
0.993366240468579 0.0283318952655477
0.993366240468579 0.024894496781619
0.993366240468579 0.0220516793054941
0.993366240468579 0.0193565883838421
0.993366240468579 0.01658653174798
0.993366240468579 0.0137503288936971
0.993366240468579 0.0110155502409928
0.993366240468579 0.00874967478109279
0.993366240468579 0.00688729125355902
0.993366240468579 0.004998449238657
0.993366240468579 0.00324336957656135
0.993366240468579 0.00192852974595448
0.993366240468579 0.000980433948591153
0.993366240468579 0.000363804201313767
0.993366240468579 0.000124207899034445
0.993366240468579 0
};
\addlegendentry{FFNN, 0.01\% }
\end{axis}

\end{tikzpicture}

%% file: figs/ensemble_sophos.tex
\begin{tikzpicture}

\definecolor{color0}{rgb}{0.603921568627451,0.803921568627451,0.196078431372549}
\definecolor{color1}{rgb}{0.419607843137255,0.556862745098039,0.137254901960784}
\definecolor{color2}{rgb}{1,0.388235294117647,0.27843137254902}
\definecolor{color3}{rgb}{0.698039215686274,0.133333333333333,0.133333333333333}

\begin{axis}[
legend cell align={left},
legend style={
  fill opacity=0.8,
  draw opacity=1,
  text opacity=1,
  at={(0.97,0.03)},
  anchor=south east,
  draw=white!80!black,
  font=\tiny
},
log basis x={10},
tick align=outside,
tick pos=left,
x grid style={white!69.0196078431373!black},
xlabel={False Positive Rate (log scale)},
xmajorgrids,
xmin=1.67845308837738e-07, xmax=2.101954791561,
xmode=log,
xtick style={color=black},
y grid style={white!69.0196078431373!black},
ylabel={True Positive Rate},
ymajorgrids,
ymin=-0.0422505295372264, ymax=1.04963097759701,
ytick style={color=black},
width=\columnwidth,
height=\columnwidth,
]
\addplot [semithick, color0, dashed]
table {%
0 0.14274500926782
0 0.14274500926782
0 0.14274500926782
0 0.14274500926782
0 0.14274500926782
0 0.14274500926782
0 0.436187273173593
0 0.436187273173593
1.05840975345756e-06 0.539902338783292
1.05840975345756e-06 0.539902338783292
3.88083576267772e-06 0.631867631127528
3.88083576267772e-06 0.631666252640337
6.70326177189788e-06 0.647395823380777
8.46727802766048e-06 0.670641809407756
1.02312942834231e-05 0.695984630558671
1.6934556055321e-05 0.734967536905915
2.01097853156936e-05 0.754646036886071
2.61074405852865e-05 0.779950640221899
4.05723738825398e-05 0.80315546860186
5.96237494447759e-05 0.825934756309982
8.71424030346724e-05 0.843747932930674
0.000116425072880332 0.859527480813922
0.000161231085776702 0.873529165337618
0.000226146883988765 0.885493546333956
0.000306233222000387 0.897196282288542
0.000429714359903769 0.908034707655763
0.00061458326350769 0.918969412518686
0.000834732492226862 0.927096577888642
0.0011695427775706 0.934202886620972
0.0015745609098937 0.939674648800328
0.00215456945478844 0.945437454340735
0.0029335590333332 0.951869071645174
0.00390694320326301 0.960175566762848
0.00523665865685685 0.967232633310354
0.0071534387203685 0.971593873978225
0.00969750296442932 0.974828424058997
0.0132763391441205 0.977597010778894
0.0182152318570046 0.980195822204845
0.0249093207443725 0.982773319849304
0.0342201513455387 0.985516183039816
0.0474051144476107 0.988278890095853
0.0654143092059422 0.990670443370752
0.0907533443102185 0.992606322696531
0.125950407858198 0.994423873787136
0.175503388498826 0.99628037765081
0.24453675345509 0.997727509918258
0.339179753609265 0.998790259160884
0.475117668704341 0.99950831310974
0.678088907124897 0.999854478319474
0.999998941590246 1
};
\addlegendentry{Single FFNN}
\addplot [semithick, color0, dashed, forget plot]
table {%
3.17522926037268e-06 0.520323793088749
3.17522926037268e-06 0.520323793088749
3.17522926037268e-06 0.520323793088749
3.17522926037268e-06 0.520323793088749
3.17522926037268e-06 0.520323793088749
3.17522926037268e-06 0.520323793088749
3.5280325115252e-06 0.531872922825002
3.5280325115252e-06 0.531872922825002
3.5280325115252e-06 0.538791082313824
3.5280325115252e-06 0.538791082313824
4.93924551613528e-06 0.579834810843864
5.64485201844032e-06 0.589459820582057
6.70326177189788e-06 0.605598027960741
8.820081278813e-06 0.648007308422177
1.02312942834231e-05 0.666115938151816
1.48177365484058e-05 0.68924359594362
1.97569820645411e-05 0.718938103308634
3.13994893525743e-05 0.750996970503196
4.76284389055902e-05 0.77769431921577
8.50255835277573e-05 0.813391963381453
0.000124892350907992 0.835095272603265
0.000177107232078565 0.854346027037634
0.000250137505067137 0.871980608868591
0.000346099989380622 0.886745179778072
0.000450529751721768 0.898749983463445
0.00057259967662054 0.908053816563307
0.000744062056680665 0.91688139689054
0.00100266683977546 0.925233459403126
0.00129549353823205 0.931798839060371
0.00172944153714965 0.938502390818317
0.00230980288529555 0.945724087953892
0.00310396300363987 0.954492871642528
0.00416060874084167 0.961246400543281
0.0056286230688873 0.96621545146264
0.00758526989977918 0.970193779021653
0.0103727683871352 0.9736304425476
0.0141184805046215 0.97653793632618
0.0192380084820958 0.978906705903623
0.0262400946077198 0.981198304892909
0.0357534342750475 0.983420817831845
0.0490086052240989 0.985892481526831
0.0669116062038335 0.988786746061728
0.0910874489890599 0.992331448411094
0.125086039892875 0.99535433059292
0.17117025896817 0.997096915969314
0.234336153054518 0.998145701010273
0.323402745020976 0.998881393950708
0.454030618382954 0.999451721345091
0.66210445022493 0.999862562857281
1 1
};
\addplot [semithick, color0, dashed, forget plot]
table {%
1.05840975345756e-06 0.533929335259903
1.05840975345756e-06 0.533929335259903
1.05840975345756e-06 0.533929335259903
1.05840975345756e-06 0.533929335259903
1.05840975345756e-06 0.533929335259903
1.05840975345756e-06 0.533929335259903
1.41121300461008e-06 0.552453216249627
1.41121300461008e-06 0.552453216249627
1.7640162557626e-06 0.585086085628485
1.7640162557626e-06 0.585086085628485
2.11681950691512e-06 0.606362384262492
3.5280325115252e-06 0.626866242056942
3.88083576267772e-06 0.635022070788213
4.23363901383024e-06 0.669166748736975
5.2920487672878e-06 0.696643887868931
6.70326177189788e-06 0.717249904822941
1.02312942834231e-05 0.739516926817294
1.41121300461008e-05 0.759733416040605
2.29322113249138e-05 0.797133222893647
3.56331283664045e-05 0.820811364214308
5.11564714171154e-05 0.840640530580867
7.51470924954868e-05 0.85737184905139
0.000112544237117654 0.877299499787597
0.000166170331292837 0.893215014897598
0.000246962275806764 0.905329327322357
0.000347511202385232 0.914069447649678
0.000493571748362375 0.922768410329981
0.00071125135432348 0.93012313486038
0.000996669184505869 0.936552547290871
0.00143696764194421 0.942015489974438
0.00198910472999791 0.947319681733795
0.00271023457535366 0.954143031642881
0.00372454392241715 0.962542131466344
0.00501227578912385 0.968222621712717
0.00676112150508689 0.971845229608223
0.00913584018859451 0.974645419521366
0.0124740645509997 0.977154566073458
0.0171169553361668 0.9796490134659
0.0236043015183594 0.982180943715448
0.0325944339642279 0.985005387242011
0.0451499960662438 0.988073101860767
0.0621378254124887 0.991132731941715
0.0856200570059493 0.993650697989596
0.119232681153003 0.995612300844761
0.166388363702049 0.997340187061506
0.231545479337901 0.99861827899299
0.323285614341593 0.999236378656232
0.457891344360317 0.999617086891142
0.668617198241205 0.99988608151272
0.999998235983744 1
};
\addplot [semithick, color0, dashed, forget plot]
table {%
7.0560650230504e-07 0.448215595514404
7.0560650230504e-07 0.448215595514404
7.0560650230504e-07 0.448215595514404
7.0560650230504e-07 0.448215595514404
7.0560650230504e-07 0.448215595514404
7.0560650230504e-07 0.448215595514404
7.0560650230504e-07 0.476759158678898
7.0560650230504e-07 0.476759158678898
7.0560650230504e-07 0.523250395774874
7.0560650230504e-07 0.523250395774874
2.11681950691512e-06 0.603578363424963
2.46962275806764e-06 0.613535574171225
4.58644226498276e-06 0.631854401883844
5.2920487672878e-06 0.652503046400837
8.46727802766048e-06 0.674381275622473
1.44649332972533e-05 0.725715885822808
2.04625885668462e-05 0.751231422099599
3.13994893525743e-05 0.785983910299848
4.05723738825398e-05 0.808904309940601
5.2920487672878e-05 0.824877151773233
6.77382242212838e-05 0.839990827724379
9.52568778111804e-05 0.858988021654802
0.000132301219182195 0.877678738106542
0.000183810493850463 0.891881066159448
0.000248373488811374 0.903790325307102
0.0003499808251433 0.914168666977309
0.000498863797129663 0.92256115217893
0.000718660222597683 0.929368333012402
0.000991024332487429 0.935022364771406
0.00137487426974137 0.940264085102255
0.00189878709770286 0.946130519718188
0.00258640063419912 0.95461928441551
0.00355837359112432 0.963457889112479
0.00491878292756843 0.968971543896835
0.00669620570687483 0.97257871767471
0.00943219491956262 0.975618503890133
0.013226593885708 0.978317269601697
0.0180099003648338 0.980936659851156
0.0243335458384916 0.983667028755966
0.0333945917378418 0.986627439509283
0.0455278483482281 0.989182888414269
0.0624892174506367 0.991215782193732
0.0865380510654482 0.992998790259161
0.12016831537506 0.994815606391783
0.166693185711045 0.996525853616949
0.231191970480246 0.99792006890966
0.321127164051042 0.998993842522023
0.453563154075177 0.99962223159702
0.663051374151023 0.999900045714387
0.999998235983744 1
};
\addplot [semithick, color0, dashed, forget plot]
table {%
7.0560650230504e-07 0.427536082762148
7.0560650230504e-07 0.427536082762148
7.0560650230504e-07 0.427536082762148
7.0560650230504e-07 0.427536082762148
7.0560650230504e-07 0.427536082762148
7.0560650230504e-07 0.427536082762148
1.41121300461008e-06 0.498328705547904
1.41121300461008e-06 0.498328705547904
1.41121300461008e-06 0.507147466379347
1.41121300461008e-06 0.507147466379347
1.7640162557626e-06 0.550345356755954
2.11681950691512e-06 0.570106906988127
3.17522926037268e-06 0.645153466576316
3.88083576267772e-06 0.674390830076245
5.2920487672878e-06 0.693539425351053
9.52568778111804e-06 0.744031773703497
1.27009170414907e-05 0.762725429987168
2.15209983203037e-05 0.790910333656225
3.31635056083369e-05 0.810734355316907
4.55116193986751e-05 0.825503335974282
6.35045852074536e-05 0.83724575965992
9.13760420485027e-05 0.851249649057563
0.000132301219182195 0.867692863998965
0.000197217017394259 0.88301600297511
0.000286123436684694 0.895779283298374
0.000408193361583466 0.906626528161385
0.000576127709132065 0.915492326303705
0.000792043298837407 0.923599647808135
0.00108592840704746 0.930684642758973
0.0015195236027139 0.937254432164113
0.00205684295421919 0.943370017536098
0.00276738870204037 0.951359745763335
0.00370619815335722 0.959909511973201
0.00504649770448565 0.965490782891942
0.00680239948547174 0.970152621374636
0.00946641683492442 0.974171371622684
0.0132579933750605 0.977318461703544
0.0181841851709032 0.979878320356425
0.0246478935352685 0.981830368757818
0.033297923647026 0.983606762201405
0.0453373345926057 0.985258947745957
0.0621618160335671 0.986826613122528
0.0860709395609222 0.988234792616906
0.119591482059425 0.989871544043827
0.166556298049598 0.992274121688463
0.230356179578266 0.99585483697897
0.318499838239709 0.998649882186235
0.448496546585376 0.999599447899564
0.657098172091075 0.99991621479
0.999998941590246 1
};
\addplot [ultra thick, color1]
table {%
0 0.59617439670974
0 0.59617439670974
0 0.59617439670974
0 0.59617439670974
0 0.59617439670974
0 0.59617439670974
7.0560650230504e-07 0.668115758822068
7.0560650230504e-07 0.668115758822068
7.0560650230504e-07 0.697392810053049
7.0560650230504e-07 0.697392810053049
1.05840975345756e-06 0.724927275907636
1.41121300461008e-06 0.747483136389093
2.46962275806764e-06 0.77566510022622
4.23363901383024e-06 0.797453664573996
5.99765526959284e-06 0.814225405733554
9.17288452996552e-06 0.833470280504064
1.19953105391857e-05 0.838133588902722
2.15209983203037e-05 0.86139427408935
3.4574718612947e-05 0.878993577937149
5.15092746682679e-05 0.889648998766741
6.9855043728199e-05 0.896057097415741
9.7726500569248e-05 0.904168093710083
0.00013724046469833 0.910998793198993
0.000194394591385039 0.9181521392422
0.000262485618857475 0.924626384109621
0.000377499478733196 0.930960252002393
0.000568013234355557 0.936497425442188
0.00078675125007012 0.941038730815759
0.00108839802980552 0.946045264592223
0.00149764980114245 0.955886351977257
0.00205543174121458 0.962628121550291
0.00281043069868097 0.967660378856141
0.0038430858148044 0.971734250952873
0.00512270320673459 0.974282350278035
0.00693928714691892 0.976449741368286
0.00958919236632549 0.978489249769591
0.0133673623829178 0.980332524389581
0.0185316963732884 0.982127291782729
0.0253076356149237 0.983749344050001
0.0344724056701127 0.986040943039286
0.0469831617592322 0.989135851103392
0.0645566445023904 0.992230759167498
0.088637936016308 0.994406969753539
0.12242449216618 0.996449417986774
0.171214359374565 0.99819273832115
0.236550346258751 0.998948275127111
0.328166647321288 0.999428202689652
0.462675356445945 0.99976628336158
0.674259580636888 0.999943408235351
0.999999647196749 1
};
\addlegendentry{FFNN Ensemble}
\addplot [semithick, color2, dashed]
table {%
0 0.0389909908850511
0 0.0389909908850511
0 0.0389909908850511
0 0.0389909908850511
0 0.0389909908850511
0 0.0389909908850511
2.11681950691512e-06 0.0652627989257854
2.11681950691512e-06 0.0652627989257854
2.11681950691512e-06 0.0760865251333581
2.11681950691512e-06 0.0760865251333581
2.11681950691512e-06 0.0805095022717551
4.93924551613528e-06 0.0950285972150972
5.2920487672878e-06 0.097305497044734
5.99765526959284e-06 0.0975090804058732
7.76167152535544e-06 0.103673173004699
9.87849103227056e-06 0.111992162408075
1.41121300461008e-05 0.126082041889665
2.04625885668462e-05 0.153100567240571
2.78714568410491e-05 0.176650825872285
4.12779803848448e-05 0.205561133070022
5.32732909240305e-05 0.233231566151363
7.47942892443342e-05 0.262704116205677
9.80793038204006e-05 0.283927497864947
0.000128420383419517 0.301508427763185
0.000186632919859683 0.320835617827729
0.000307291631753845 0.35156935577993
0.000426186327392244 0.37894580566829
0.0006159944765123 0.419701430669209
0.000905645945708519 0.471815096330943
0.00130501922601317 0.51678129561333
0.00246785874181188 0.561478500274139
0.00438393319882121 0.612366256021143
0.00622944700560005 0.665013501178138
0.0083646122815751 0.713328903986559
0.0108416439079169 0.731131791195497
0.0144751645915367 0.740728872530357
0.0190143312208651 0.784897642401784
0.024431625142312 0.819809616484226
0.0311945106636547 0.84865304250556
0.0415993841466448 0.890206831875422
0.0556074372336556 0.926920187972854
0.0753150268430354 0.950136775680534
0.10222685884095 0.959666975838991
0.137728391594674 0.97269557599392
0.185375881875827 0.981269595817207
0.247595910445834 0.987157344214631
0.335454151277095 0.991087164546803
0.454199258337005 0.995185290256956
0.634885679398513 0.999480384706406
0.999999294393498 1
};
\addlegendentry{Single LGBM}
\addplot [semithick, color2, dashed, forget plot]
table {%
1.7640162557626e-06 0.0236619722450468
1.7640162557626e-06 0.0236619722450468
1.7640162557626e-06 0.0236619722450468
1.7640162557626e-06 0.0236619722450468
1.7640162557626e-06 0.0236619722450468
1.7640162557626e-06 0.0236619722450468
5.99765526959284e-06 0.0449801634840536
5.99765526959284e-06 0.0449801634840536
5.99765526959284e-06 0.045887836592382
5.99765526959284e-06 0.045887836592382
6.70326177189788e-06 0.0522547775943649
8.11447477650796e-06 0.0637561350617585
8.46727802766048e-06 0.0714709890035587
1.02312942834231e-05 0.0755779342095012
1.27009170414907e-05 0.0913346983952927
1.55233430507109e-05 0.1026302676276
1.65817528041684e-05 0.110552379720451
2.29322113249138e-05 0.127213877182641
2.6460243836439e-05 0.150857475478127
2.96354730968117e-05 0.166291593109622
4.05723738825398e-05 0.188441756784765
5.15092746682679e-05 0.19830415795129
7.62055022489443e-05 0.225458650528949
0.000102312942834231 0.249994487815132
0.00013582925169372 0.278331527786556
0.000184516100352768 0.315701936320301
0.000262485618857475 0.383426109529318
0.000375382659226281 0.430600857548974
0.000601882346466199 0.475350244226538
0.00112226714191617 0.518619425527443
0.00189243663918212 0.557595717252845
0.00315970591732197 0.606994448127401
0.00457762218370395 0.653249763711009
0.00603258279145694 0.698332086354623
0.00906245711235478 0.737456104634498
0.0138221257736534 0.7433071051328
0.0181739538766198 0.789746160212021
0.0233492247677761 0.838145348230442
0.0304726752117966 0.876521914242163
0.0400110639099561 0.903866025979295
0.0528284060243272 0.932878492336593
0.0722336432474693 0.951493508116141
0.0979156031118658 0.961405886425473
0.135340971994125 0.970118813307443
0.183356083262978 0.978746485063449
0.247825232559083 0.985179572283853
0.332386527008324 0.990072922531019
0.443188268868535 0.996218641180284
0.629287750212476 0.999689112773423
0.999998235983744 1
};
\addplot [semithick, color2, dashed, forget plot]
table {%
7.0560650230504e-07 0.0152489082199171
7.0560650230504e-07 0.0152489082199171
7.0560650230504e-07 0.0152489082199171
7.0560650230504e-07 0.0152489082199171
7.0560650230504e-07 0.0152489082199171
7.0560650230504e-07 0.0152489082199171
3.5280325115252e-06 0.0675051557302469
3.5280325115252e-06 0.0675051557302469
3.88083576267772e-06 0.0708124666512815
3.88083576267772e-06 0.0708124666512815
4.93924551613528e-06 0.0764437147128299
5.99765526959284e-06 0.0801750963897394
8.11447477650796e-06 0.0874092878110158
8.46727802766048e-06 0.0883610583982914
1.05840975345756e-05 0.0963831247767565
1.30537202926432e-05 0.111629093164744
1.65817528041684e-05 0.123192187102663
2.11681950691512e-05 0.130158118860345
2.11681950691512e-05 0.130158118860345
3.4574718612947e-05 0.16381845949867
4.86868486590478e-05 0.186723425021792
7.09134534816565e-05 0.211615716929463
0.000106899385099214 0.226475832376663
0.000134770841940263 0.245736876222786
0.000167581544297447 0.260967410493142
0.00022826370349568 0.2790877995505
0.000337985514604114 0.302843111459318
0.000563073988839422 0.330065220171363
0.00100231403652431 0.36400778467495
0.00222971654728393 0.432747669815717
0.00288628339767877 0.529143288878175
0.00351109795546988 0.599553733513055
0.00464183237541371 0.666155625882868
0.00589922316252129 0.723670497757643
0.00757080496648193 0.770355028803003
0.0101169860300497 0.806551709438771
0.0136467825578306 0.838210024532898
0.0209614523639758 0.852252866703611
0.0280418608113557 0.87153081458333
0.0387021638481803 0.896314332709599
0.0517562369440747 0.918083053191849
0.0714634737502033 0.93369870544501
0.0959381408891559 0.9464818296338
0.12973457553006 0.956676431808394
0.181060392507729 0.96774122423421
0.247305906173387 0.97723614640951
0.33144242550824 0.989603284380232
0.447365812165432 0.996612578658878
0.626490726037339 0.999607532437371
0.99999400234473 1
};
\addplot [semithick, color2, dashed, forget plot]
table {%
1.05840975345756e-06 0.0483984530604385
1.05840975345756e-06 0.0483984530604385
1.05840975345756e-06 0.0483984530604385
1.05840975345756e-06 0.0483984530604385
1.05840975345756e-06 0.0483984530604385
1.05840975345756e-06 0.0483984530604385
1.05840975345756e-06 0.0601085385948485
1.05840975345756e-06 0.0601085385948485
1.41121300461008e-06 0.0846928831078727
1.41121300461008e-06 0.0846928831078727
1.41121300461008e-06 0.0891430536916205
1.7640162557626e-06 0.106119848128283
3.17522926037268e-06 0.121346707608726
4.58644226498276e-06 0.139131220868103
6.35045852074536e-06 0.143661501871938
8.46727802766048e-06 0.152730883375397
1.48177365484058e-05 0.180651937128754
2.32850145760663e-05 0.199519778454266
3.66915381198621e-05 0.223849092547379
4.62172259009801e-05 0.243457771519202
6.35045852074536e-05 0.266876472672057
7.09134534816565e-05 0.276112689637533
0.000102312942834231 0.301180636503011
0.00013582925169372 0.332615524370472
0.000196158607640801 0.370566549710353
0.000256487963587882 0.412252631517056
0.000371854626714756 0.458148552647245
0.000641396310595281 0.49773118470817
0.00124363146031263 0.537288093239709
0.00265025802265773 0.577918775383611
0.00389353667971921 0.620701414500133
0.00478895133114431 0.648676855144191
0.00580220226845434 0.684083455948823
0.00688601385599489 0.722676099607385
0.00896543621828784 0.754221231172214
0.0114886850705307 0.786734302399932
0.0153257732300655 0.824512612613937
0.0228528305934045 0.839691699825521
0.0303502524836467 0.86719088769695
0.0400195311879838 0.89591010581925
0.0556660025733469 0.921028764785517
0.0738226690906602 0.935525076031403
0.100233873275189 0.95082469635211
0.136660103350184 0.963188894490902
0.182840990516296 0.972121573809625
0.246148358706355 0.981804645228432
0.329427213337656 0.990239757993036
0.439857453374404 0.99601358790318
0.624855835771498 0.999681763193598
0.999998235983744 1
};
\addplot [semithick, color2, dashed, forget plot]
table {%
3.5280325115252e-07 0.00738044805978442
3.5280325115252e-07 0.00738044805978442
3.5280325115252e-07 0.00738044805978442
3.5280325115252e-07 0.00738044805978442
3.5280325115252e-07 0.00738044805978442
3.5280325115252e-07 0.00738044805978442
3.5280325115252e-07 0.0280136584591459
3.5280325115252e-07 0.0280136584591459
2.11681950691512e-06 0.0345378804693736
2.11681950691512e-06 0.0345378804693736
3.5280325115252e-06 0.0375556179453221
3.5280325115252e-06 0.0373689386177792
4.58644226498276e-06 0.0403572777744296
9.17288452996552e-06 0.0529044804508526
1.16425072880332e-05 0.0593537367468702
1.34065235437958e-05 0.0637135074987763
1.55233430507109e-05 0.0712475617768932
2.22266048226088e-05 0.0992207975470042
2.7165850338744e-05 0.115932272152001
3.13994893525743e-05 0.123561136009854
4.58644226498276e-05 0.135695292300139
6.35045852074536e-05 0.155448023036523
7.9380731509317e-05 0.175130932764574
0.000115366663126874 0.206602568531157
0.000152763807749041 0.229895591869013
0.000216268392956495 0.265386712841627
0.000353508857654825 0.301515777343009
0.000650921998376399 0.333063113781785
0.00108204757128478 0.377203220291896
0.00132689302758463 0.438878689305332
0.00183034326697927 0.492177842192762
0.00241352704113439 0.548429321295702
0.00371537103788719 0.600221075361122
0.00657448858522721 0.651499093796808
0.00898766282311045 0.692380396612726
0.0111097743787929 0.726659571872276
0.0147542319631984 0.76433572292672
0.0201764651301615 0.799843012974948
0.0295313961377217 0.819375256316596
0.0403758624716479 0.861615496442068
0.056316218965221 0.894291728341891
0.0753781786249917 0.918964267812809
0.101036853474812 0.942621830309961
0.137948893626645 0.960401198863461
0.185985525893818 0.972176695658309
0.249918414248171 0.98237350270685
0.336249369805193 0.99043599177435
0.451858055962357 0.9960797341216
0.626838590042975 0.999559025210529
0.999996471967488 1
};
\addplot [ultra thick, color3]
table {%
1.41121300461008e-06 0.075605127654852
1.41121300461008e-06 0.075605127654852
1.41121300461008e-06 0.075605127654852
1.41121300461008e-06 0.075605127654852
1.41121300461008e-06 0.075605127654852
1.41121300461008e-06 0.075605127654852
1.41121300461008e-06 0.079182903113429
1.41121300461008e-06 0.079182903113429
1.7640162557626e-06 0.0821102407575359
1.7640162557626e-06 0.0821102407575359
2.82242600922016e-06 0.0872968392397007
2.82242600922016e-06 0.0872167288196134
2.82242600922016e-06 0.0872167288196134
3.88083576267772e-06 0.0973848725068388
5.64485201844032e-06 0.105030640398288
5.99765526959284e-06 0.119719510635577
7.40886827420292e-06 0.129327616340174
1.79929658087785e-05 0.157012013623181
2.85770633433541e-05 0.194544112913065
3.5280325115252e-05 0.202815330047581
3.88083576267772e-05 0.209005881133776
5.00980616636578e-05 0.246772432020061
7.51470924954868e-05 0.297591836674697
0.000109369007857281 0.339994502514291
0.000152058201246736 0.371645468028593
0.000219090818965715 0.408760846142426
0.000300235566730795 0.448932179547295
0.000713368173830396 0.495524105886866
0.00199404397551404 0.545285906004754
0.002597690338236 0.585013324788222
0.00326554689266773 0.635574759191017
0.00388295258218464 0.686203809728198
0.00480800270670654 0.732876581445839
0.00640090938566017 0.777882468459278
0.00929742407762236 0.803153263727913
0.0131884911345835 0.807060300362628
0.0179721504169605 0.830885433279779
0.0236808598238594 0.867845735259315
0.0300500169169159 0.894385068005662
0.040043169005811 0.916558015378261
0.0546121792621543 0.93807758510446
0.0735009125256091 0.949542929630713
0.0979410049459488 0.959971983401709
0.137477195679854 0.96954848591306
0.187085566430912 0.980284752120721
0.250956714216313 0.985550726064991
0.333226551549318 0.991414220848994
0.445799012927064 0.996056215466162
0.625004718743484 0.999712631428861
0.999994707951233 1
};
\addlegendentry{LGBM Ensemble}
\end{axis}

\end{tikzpicture}